\documentclass[twoside,11pt]{article}

\usepackage{jmlr2e}

\usepackage{hyperref}       
\usepackage{url}               
\usepackage{amsfonts}     

\usepackage{amsmath}
\usepackage{algorithm}
\usepackage{algorithmic}
\usepackage{color}
\usepackage{graphicx}

\newcommand{\A}{\ensuremath{\mathbf{A}}}
\newcommand{\B}{\ensuremath{\mathbf{B}}}
\newcommand{\C}{\ensuremath{\mathbf{C}}}
\newcommand{\D}{\ensuremath{\mathbf{D}}}
\newcommand{\E}{\ensuremath{\mathbf{E}}}

\newcommand{\HH}{\ensuremath{\mathbf{H}}}
\newcommand{\I}{\ensuremath{\mathbf{I}}}

\newcommand{\M}{\ensuremath{\mathbf{M}}}

\newcommand{\PP}{\ensuremath{\mathbf{P}}}
\newcommand{\Q}{\ensuremath{\mathbf{Q}}}

\newcommand{\T}{\ensuremath{\mathbf{T}}}
\newcommand{\U}{\ensuremath{\mathbf{U}}}
\newcommand{\V}{\ensuremath{\mathbf{V}}}

\newcommand{\X}{\ensuremath{\mathbf{X}}}

\newcommand{\Z}{\ensuremath{\mathbf{Z}}}
\renewcommand{\aa}{\ensuremath{\mathbf{a}}}
\renewcommand{\b}{\ensuremath{\mathbf{b}}}
\renewcommand{\c}{\ensuremath{\mathbf{c}}}

\newcommand{\f}{\ensuremath{\mathbf{f}}}
\newcommand{\g}{\ensuremath{\mathbf{g}}}

\newcommand{\m}{\ensuremath{\mathbf{m}}}

\newcommand{\p}{\ensuremath{\mathbf{p}}}
\newcommand{\q}{\ensuremath{\mathbf{q}}}
\newcommand{\rr}{\ensuremath{\mathbf{r}}}

\newcommand{\uu}{\ensuremath{\mathbf{u}}}
\newcommand{\vv}{\ensuremath{\mathbf{v}}}
\newcommand{\w}{\ensuremath{\mathbf{w}}}
\newcommand{\x}{\ensuremath{\mathbf{x}}}
\newcommand{\y}{\ensuremath{\mathbf{y}}}
\newcommand{\z}{\ensuremath{\mathbf{z}}}
\newcommand{\0}{\ensuremath{\mathbf{0}}}

\newcommand{\bphi}{\ensuremath{\boldsymbol{\phi}}}

\newcommand{\bpsi}{\ensuremath{\boldsymbol{\psi}}}

\newcommand{\bPhi}{\ensuremath{\boldsymbol{\Phi}}}

\newcommand{\bSigma}{\ensuremath{\boldsymbol{\Sigma}}}
\newcommand{\bTheta}{\ensuremath{\boldsymbol{\Theta}}}
\newcommand{\bXi}{\ensuremath{\boldsymbol{\Xi}}}

\newcommand{\bbE}{\ensuremath{\mathbb{E}}}

\newcommand{\bbP}{\ensuremath{\mathbb{P}}}
\newcommand{\bbR}{\ensuremath{\mathbb{R}}}

\newcommand{\calE}{\ensuremath{\mathcal{E}}}
\newcommand{\calF}{\ensuremath{\mathcal{F}}}

\newcommand{\calN}{\ensuremath{\mathcal{N}}}
\newcommand{\calO}{\ensuremath{\mathcal{O}}}

\newcommand{\abs}[1]{\left\lvert#1\right\rvert}
\newcommand{\norm}[1]{\left\lVert#1\right\rVert}

\newcommand{\caja}[4][1]{{%
    \renewcommand{\arraystretch}{#1}%
    \begin{tabular}[#2]{@{}#3@{}}%
      #4%
    \end{tabular}%
    }}

{%
\begin{list}{#1}{
\vspace{-\topsep}
\vspace{-\partopsep}
\setlength{\itemindent}{0cm}
\setlength{\rightmargin}{0cm}
\setlength{\listparindent}{0cm}
\settowidth{\labelwidth}{#1}
\setlength{\leftmargin}{\labelwidth}
\addtolength{\leftmargin}{\labelsep}
\setlength{\itemsep}{0cm}
}%
}%
{%
\end{list}
\vspace{-\topsep}
\vspace{-\partopsep}
}

{\begin{enumerate}%
}%
{\end{enumerate}}

\newcommand{\diagop}{\operatorname{diag}}
\newcommand{\diag}[1]{\ensuremath{\diagop\left(#1\right)}}
\newcommand{\traceop}{\operatorname{tr}}
\newcommand{\trace}[1]{\ensuremath{\traceop\left(#1\right)}}
\newcommand{\rankop}{\operatorname{rank}}
\newcommand{\rank}[1]{\ensuremath{\rankop\left(#1\right)}}

\DeclareMathOperator*{\argmin}{arg\,min}

\newcommand{\ie}{i.e.\@}
\newcommand{\eg}{e.g.\@}
\newcommand{\Sxy}{\bSigma_{xy}}
\newcommand{\Sxx}{\bSigma_{xx}}
\newcommand{\Syy}{\bSigma_{yy}}
\newcommand{\tu}{\tilde{\uu}}

\newcommand{\tw}{\tilde{\w}}

\newcommand{\Exy}{\E_{xy}}
\newcommand{\Exx}{\E_{xx}}
\newcommand{\Eyy}{\E_{yy}}
\newcommand{\normhatM}[1]{\norm{#1}_{\widehat{\M}_{\lambda}^{-1}}}

\newcommand{\smallnorm}[1]{\|#1\|}

\newcommand{\rbr}[1]{\left(#1\right)}
\newcommand{\sbr}[1]{\left[#1\right]}

\newcommand{\inner}[2]{\left\langle #1,#2 \right\rangle}

\usepackage{mathtools}
\DeclarePairedDelimiter{\ceil}{\lceil}{\rceil}

\usepackage{lastpage}

\jmlrheading{20}{2019}{1-\pageref{LastPage}}{2/18}{10/19}{18-095}{Chao Gao, Dan Garber,
  Nathan Srebro, Jialei Wang, and Weiran Wang}

\ShortHeadings{Stochastic Canonical Correlation Analysis}{Gao, Garber,
  Srebro, Wang, and Wang}
\firstpageno{1}

\begin{document}

\title{Stochastic Canonical Correlation Analysis}

\author{\name{Chao Gao} \email{chaogao@galton.uchicago.edu} \\
\addr University of Chicago \\
Chicago, IL 60637, USA
\AND
\name{Dan Garber} \email{dangar@technion.ac.il} \\
\addr Technion -- Israel Institute of Technology \\
Haifa, 3200003, Israel
\AND
\name{Nathan Srebro} \email{nati@ttic.edu} \\
\addr Toyota Technological Institute at Chicago \\
Chicago, IL 60637, USA
\AND
\name{Jialei Wang} \email{jialei@uchicago.edu} \\
\addr University of Chicago \\
Chicago, IL 60637, USA
\AND
\name{Weiran Wang} \email{weiranwang@ttic.edu} \\
\addr Toyota Technological Institute at Chicago \\
Chicago, IL 60637, USA
}

\editor{John Shawe-Taylor}

\maketitle

\begin{abstract}
  We study the sample complexity of canonical correlation analysis (CCA), \ie, the number of samples needed to estimate the population canonical correlation and directions up to arbitrarily small error. With mild assumptions on the data distribution, we show that in order to  achieve $\epsilon$-suboptimality in a properly defined measure of alignment between the estimated canonical directions and the population solution, we can solve the empirical objective exactly with $N(\epsilon, \Delta, \gamma)$ samples, where $\Delta$ is the singular value gap of the whitened cross-covariance matrix and $1/\gamma$ is an upper bound of the condition number of auto-covariance matrices. 
Moreover, we can achieve the same learning accuracy by drawing the same
level of samples and solving the empirical objective approximately with a
stochastic optimization algorithm; this algorithm is based on the
shift-and-invert power iterations and only needs to process the dataset
for $\calO\left(\log \frac{1}{\epsilon} \right)$ passes. Finally, we show
that, given an estimate of the canonical correlation, the streaming
version of the shift-and-invert power iterations achieves the same learning accuracy with the same level of sample complexity, by processing the data only once.
\end{abstract}

\section{Introduction}
\label{sec:intro}

Let $\x \in \bbR^{d_x}$ and $\y \in \bbR^{d_y}$ be two random vectors with a joint probability distribution $P(\x,\y)$. The objective 
of CCA~\citep{Hotell36a} in the population setting is to find 
$\uu \in \bbR^{d_x}$ and $\vv \in \bbR^{d_y}$ such that projections of the random variables onto these directions are maximally correlated:\footnote{For simplicity (especially for the streaming setting), we assume that $\bbE [\x] = \0$ and $\bbE [\y] = \0$. Nonzero means can be easily handled in the ERM approach (see Remark~\ref{rmk:erm-handle-means}).}
\begin{align} \label{e:cca-ratio}
\max_{\uu,\vv}\; \frac{\bbE[ (\uu^\top \x) (\vv^\top \y) ]}{\sqrt{ \bbE[(\uu^\top \x)^2]} \sqrt{ \bbE[(\vv^\top \y)^2]}}.
\end{align}
This objective can be written in the equivalent constrained form
\begin{gather} \label{e:cca-population}
  \max_{\uu,\vv} \; \uu^\top \Exy \vv  \quad  \text{s.t.} \quad \uu^\top \Exx \uu = \vv^\top \Eyy \vv = 1 
\end{gather}
where the cross- and auto-covariance matrices are defined as 
\begin{gather} \label{e:cov-population}
\Exy = \bbE [ \x \y^\top],  \qquad \Exx = \bbE [ \x \x^\top],  \qquad \Eyy = \bbE [ \y \y^\top].
\end{gather} 
The global optimum of~\eqref{e:cca-population}, denoted by $(\uu^*,\vv^*)$, can be computed in closed-form. Define 
\begin{align} \label{e:T-population}
  \T:=\Exx^{-\frac{1}{2}} \Exy \Eyy^{-\frac{1}{2}} \ \in \bbR^{d_x \times d_y},
\end{align}
and let $(\aa_1,\b_1)$ be the (unit-length) top left and right singular vector pair associated with $\T$'s largest singular value $\rho_1 = \sigma_1 (\T)$. Then the optimal objective value, \ie, the canonical correlation between $\x$ and $\y$, is $\rho_1 \le 1$ (see Lemma~\ref{lem:population-obj-bound}), 
achieved by $(\uu^*,\,\vv^*)=(\Exx^{-\frac{1}{2}} \aa_1,\,\Eyy^{-\frac{1}{2}} \b_1 )$. 

In practice, we do not have access to the population covariance matrices, but observe samples pairs ${(\x_1,\y_1), \dots, (\x_N,\y_N)}$ drawn from $P(\x,\y)$. In this paper, we are concerned with both the number of samples $N(\epsilon)$ needed to approximately solve~\eqref{e:cca-population}, and the time complexity for obtaining the approximate solution. Note that the CCA objective is not a stochastic convex program due to the ratio form~\eqref{e:cca-ratio}, and standard stochastic approximation methods do not apply~\citep{Arora_12a}. Globally convergent stochastic optimization of CCA has long been a challenge even for the empirical objective, and attracted continuous  effort~\citep{LuFoster14a,Ma_15b,WangLivesc16a}, 
until the recent breakthrough by~\citet{Ge_16a,Wang_16b}. 
And our understanding of the stochastic objective, \eg, the existence of an efficient algorithm and the sample complexity, has been very limited. 

\paragraph{Our contributions} The contributions of our paper are summarized as follows.

\begin{itemize}
\item 
First, we provide the ERM sample complexity of CCA. We show that in order to achieve $\epsilon$-suboptimality in the alignment between the estimated canonical directions and the population solution (relative to the population covariances, see Section~\ref{sec:setup}), we can solve the empirical objective exactly with $N(\epsilon, \Delta, \gamma)$ samples where $\Delta$ is the singular value gap of the whitened cross-covariance and $1/\gamma$ is a upper bound of the condition number of the auto-covariance, for several general classes of distributions widely used in statistics and machine learning. 

\item 
Second, to alleviate the high computational complexity of exactly solving the empirical objective, we show that we can achieve the same learning accuracy by drawing the same level of samples and solving the empirical objective approximately with the stochastic optimization algorithm of~\citet{Wang_16b}. This algorithm is based on the shift-and-invert power iterations~\citep{Saad92a, GarberHazan15c, Garber_16a}. We provide tightened analysis of the algorithm's time complexity, removing an extra $\log  \frac{1}{\epsilon}$ factor from the complexity given by~\citet{Wang_16b}.  Our analysis shows that asymptotically it suffices to process the sample set for $\calO\left(\log \frac{1}{\epsilon} \right)$ passes. While near-linear runtime in the required number of samples is known and achieved for convex learning problems using SGD, no such result was estabilished for the nonconvex CCA objective previously.
 
\item 
Third, we show that the streaming version of shift-and-invert power iterations achieves the same learning accuracy with the same level of sample complexity, given a good estimate of the canonical correlation. This approach requires only $\calO(d)$ memory where $d:=d_x+d_y$ is the input dimensionality, and thus further alleviates the memory cost of solving the empirical objective. This addresses the challenge of the existence of a stochastic algorithm for CCA proposed by~\citet{Arora_12a}.
\end{itemize}

\textbf{Notation\ }  We use $\sigma_i (\A)$ to denote the $i$-th largest singular value of a matrix $\A$, and use $\sigma_{\max} (\A)$ and $\sigma_{\min} (\A)$ to denote the largest and smallest singular values of $\A$ respectively. We use $\norm{\cdot}$ to denote the spectral norm of a matrix or the $\ell_2$-norm of a vector. For a positive definite matrix $\M$, the vector norm $\norm{\cdot}_{\M}$ is defined as $\norm{\w}_{\M} = \sqrt{ \w^\top \M \w }$ for any $\w$. We use $C$ and $C^\prime$ to denote universal constants that are independent of problem parameters, and their specific values may vary among appearances. We hide poly-logarithmic dependencies in the notation $\tilde{\calO} (\cdot)$.

\section{Problem setup}
\label{sec:setup}

\paragraph{Assumptions} We assume the following properties of the input random variables.
\begin{enumerate}

\item \label{assump:bounded-covariances} \textbf{Bounded covariances}: Eigenvalues of population auto-covariance matrices are bounded:\footnote{CCA is invariant to linear transformations of the inputs, so we could always rescale the data.}  
\begin{gather*}
\max \left(\norm{\Exx},\, \norm{\Eyy} \right) \le 1, \\ 
\gamma:=\min \left( \sigma_{\min} (\Exx), \sigma_{\min} (\Eyy) \right) > 0.
\end{gather*}
Hence $\Exx$ and $\Eyy$ are invertible with condition numbers bounded by $1/\gamma$.

\item \label{assump:singular-value-gap} \textbf{Singular value gap}: For the purpose of learning the canonical directions $(\uu^*,\vv^*)$, we assume that there exists a positive singular value gap $\Delta:= \sigma_1 (\T) - \sigma_2 (\T) \in (0,1)$, such that the top left- and right-singular vector pair of $\T$ is uniquely defined.
\end{enumerate}

\paragraph{Distribution classes}
In this paper, we analyze three input distribution classes commonly used in the statistics and machine learning literature. Let 
\begin{align} \label{e:distributions-z}
\z = \left[\begin{array}{cc}\Exx & \Exy \\ \Exy^\top & \Eyy\end{array}\right]^{-\frac{1}{2}} \cdot \left[ \begin{array}{c}\x \\ \y \end{array}\right] \in \bbR^d,
\end{align}
the distribution classes are defined with $(\x,\y,\z)$ as follows.
\begin{itemize}
\item \textbf{(Sub-Gaussian)} 
Let $\z$ be isotropic and sub-Gaussian, that is, $\bbE \left[ \z\z^\top \right] = \I$ and there exists constant $C>0$ such that $\bbP \left( \abs{\q^\top \z} > t \right) \le \exp (-C t^2)$ for any unit vector $\q$.  
\item \textbf{(Regular polynomial-tail,~\citealp{SrivasVershy13a})}
Let $\z$ be isotropic and regular polynomial-tail, that is, $\bbE \left[
  \z\z^\top \right] = \I$ and there exist constants $r>1, C >0$ such that
$\bbP \left( \norm{\V \z}^2 > t \right) \le C t^{-1-r}$ for any orthogonal
projection $\V$ in $\bbR^d$ and any $t > C \cdot \rank{\V}$. Note that this
class is general and only implies the existence of a $(4+\delta)$-moment
condition for some $\delta>0$.
\item \textbf{(Bounded)} 
Let $\x$ and $\y$ be bounded and in particular $\sup \left( \norm{\x}^2, \norm{\y}^2 \right) \le 1$ (which implies $\max \left(\norm{\Exx},\, \norm{\Eyy} \right) \le 1$ as in Assumption~\ref{assump:bounded-covariances}). 
\end{itemize}
As shown later, these classes satisfy the same concentration property, allowing us to study them (and potentially other distributions) in a unified framework.

\paragraph{Measure of error} For an estimate $(\uu,\vv)$ of the optimal solution to~\eqref{e:cca-population}, 
which need not be correctly normalized (\ie, they may not satisfy the constraints of~\eqref{e:cca-population}),  we can always define $(\overline{\uu}, \overline{\vv}) := \left( \frac{{\uu}}{\smallnorm{\Exx^{\frac{1}{2}} {\uu}}}, \frac{{\vv}}{\smallnorm{\Eyy^{\frac{1}{2}} {\vv}}} \right)$ as the correctly normalized version. And we can measure the quality of these directions by the alignment (cosine of the angle) between $\left( \frac{1}{\sqrt{2}} \left[ \begin{array}{c} \Exx^{\frac{1}{2}} \overline{\uu}\\ \Eyy^{\frac{1}{2}} \overline{\vv}\end{array}\right],\, \frac{1}{\sqrt{2}} \left[ \begin{array}{c} \Exx^{\frac{1}{2}} \uu^* \\ \Eyy^{\frac{1}{2}} \vv^* \end{array}\right] \right)$, or the sum of alignment between $\left( \Exx^{\frac{1}{2}} \overline{\uu},\,  \Exx^{\frac{1}{2}} \uu^* \right)$ and alignment between $\left(\Eyy^{\frac{1}{2}} \overline{\vv},\, \Eyy^{\frac{1}{2}} \vv^* \right)$ (all vectors have unit length):
\begin{align*}
\text{align} \left( ({\uu},{\vv}); (\uu^*, \vv^*) \right) := \frac{1}{2} \left(
\frac{ {\uu}^\top \Exx \uu^*}{\smallnorm{\Exx^{\frac{1}{2}} {\uu}}} + 
\frac{ {\vv}^\top \Eyy \vv^*}{\smallnorm{\Eyy^{\frac{1}{2}} {\vv}}} \right).
\end{align*}
This measure of alignment is invariant to the lengths of ${\uu}$ and ${\vv}$, and achieves the maximum of $1$ if $({\uu}, {\vv})$ lie in the same direction as $(\uu^*,\vv^*)$. Intuitively, this measure respects the geometry imposed by the CCA constraints that the projections of each view have unit length. As we will show later, this measure is also closely related to the learning guarantee we can achieve with power iterations. Moreover, high alignment 
implies accurate estimate of the canonical correlation.

\begin{lemma} \label{lem:transform-alignment-correlation}
Let $\eta \in (0,1)$. If $\text{align} \left( ({\uu},{\vv}); (\uu^*, \vv^*) \right) \ge 1 - \frac{\eta}{8}$, then 
\begin{align*}
 \frac{\uu^\top \Exy {\vv}}{ \sqrt{\uu^\top \Exx \uu} \sqrt{\vv^\top \Eyy \vv}} \ge \rho_1 (1 - \eta).
\end{align*}
\end{lemma}
All proofs are deferred to the appendix.\section{The sample complexity of ERM}
\label{sec:erm}

One approach to address this problem is empirical risk minization (ERM): We draw $N$ samples $\{(\x_i,\y_i)\}_{i=1}^N$ from $P(\x,\y)$ and solve the empirical version of~\eqref{e:cca-population}:
\begin{gather} \label{e:cca-empirical}
  \max_{\uu,\vv} \quad \uu^\top \Sxy \vv  \quad  \text{s.t.} \quad \uu^\top \Sxx \uu = \vv^\top \Syy \vv = 1
\end{gather}
where the empirical covariance matrices are defined as 
\begin{gather} \label{e:cov-empirical}
\Sxy= \frac{1}{N} \sum\limits_{i=1}^N \x_i \y_i^\top, \quad \Sxx = \frac{1}{N} \sum\limits_{i=1}^N \x_i \x_i^\top,  \quad 
\Syy= \frac{1}{N} \sum\limits_{i=1}^N \y_i \y_i^\top .
\end{gather}
Similarly, define the empirical version of $\T$ as 
\begin{align} \label{e:T-empirical}
\widehat{\T}:=\Sxx^{-\frac{1}{2}} \Sxy \Syy^{-\frac{1}{2}} \ \in \bbR^{d_x \times d_y}.
\end{align}
We will approximate the population canonical correation and directions based on solution to the above empirical objective.


Before going to the detailed analysis, we highlight the key property that
enable us to study different input distributions in a unified manner.
In fact this property is the only place we handle the stochasticity of data in studying ERM.
\begin{proposition}[Concentration property]
\label{prop:probabilistic-property} For any $\nu>0$, with sufficiently
large sample sizes $N_0 (\nu)$, the following inequality is satisfied with
high probability by sub-Gaussian, regular polynomial-tail, and bounded random variables:~\footnote{We refrain from specifying the failure probability as it only adds additional mild dependences to our results.} 
\begin{gather}  \label{e:property-probability}
\max \left(
\smallnorm{\Exx^{-\frac{1}{2}} \Sxx \Exx^{-\frac{1}{2}} - \I }, \;
\smallnorm{\Eyy^{-\frac{1}{2}} \Syy \Eyy^{-\frac{1}{2}} - \I }, \;  \smallnorm{\Exx^{-\frac{1}{2}} (\Sxy - \Exy) \Eyy^{-\frac{1}{2}}}
\right) \le \nu. 
\end{gather}
\end{proposition}
We provide detailed bounds on $N_0 (\nu)$ for different distributions in Lemma~\ref{lem:probability}. 

\paragraph{Roadmap for this section} 
We proceed to analyze the sample complexities, eventually obtained in Theorem~\ref{thm:approx-direction-erm}. 
We first analyze the concentration property of different classes in Lemma~\ref{lem:probability}, and provide the number of samples needed to guarantee small perturbation between $\widehat{\T}$ and $\T$ in Lemma~\ref{lem:approx-error-erm}, which by the Weyl's inequality~\citep{HornJohnson86a} provides the sample complexity for learning canonical correlations (regardless of the existence of a singular value gap for $\T$). Then by the perturbation of singular vectors and after fixing the issue of normalization, we obtain guarantees for the alignment between the estimated and the optimal canonical directions. 

\subsection{Approximating the canonical correlation}
\label{sec:erm-eigen-value}

We first discuss the error of approximating $\rho_1$ by $\widehat{\rho}_1 = \sigma_1 (\widehat{\T})$. Observe that, although the empirical covariance matrices are unbiased estimates of their population counterparts, we do \emph{not} have $\bbE [ \widehat{\T} ] = \T$ due to the nonlinear operations (matrix multiplication, inverse, and square root) involved in computing $\T$. Nonetheless, we can provide approximation guarantee 
based on concentrations. 
We will separate the probabilistic property of data---the concentration
property in Proposition~\ref{prop:probabilistic-property}---from the deterministic error analysis, and we show below that it is satisfied by distributions considered here.

\begin{lemma}
\label{lem:probability}
Let Assumption~\ref{assump:bounded-covariances} hold for the random variables. 
Then the concentration property~\eqref{e:property-probability} is
satisfied with high probability, if
\begin{align*}
& N_0 (\nu) \ge C^\prime \frac{d}{\nu^2} \qquad\qquad\qquad   \text{for the sub-Gaussian class,} \\
& N_0 (\nu) \ge C^\prime \frac{d}{\nu^{2 (1+r^{-1})}} \quad\qquad \text{for the polynomial-tail class,} \\
& N_0 (\nu) \ge C \frac{1}{\nu^2 \gamma^2} \qquad\quad\qquad\;    \text{for the bounded class.}
\end{align*}
\end{lemma}

\begin{remark} \label{rmk:erm-handle-means}
When $(\x,\y)$ have nonzero means, we use the unbiased estimate of covariance matrices
$\Sxy= \frac{\sum\nolimits_{i=1}^N (\x_i - \bar{\x}) (\y_i - \bar{\y})^\top}{N-1}$,  $\Sxx= \frac{\sum\nolimits_{i=1}^N (\x_i - \bar{\x}) (\x_i - \bar{\x})^\top}{N-1}$, and
$\Syy= \frac{\sum\nolimits_{i=1}^N (\y_i - \bar{\y}) (\y_i - \bar{\y})^\top}{N-1}$
instead of those in~\eqref{e:cov-empirical}, where $\bar{\x} = \frac{1}{N} \sum_{i=1}^N \x_i$ and $\bar{\y} = \frac{1}{N} \sum_{i=1}^N \y_i$. We have similar concentration results, and all results in Sections~\ref{sec:erm} and~\ref{sec:stochastic-opt} still apply.
\end{remark}

We will decompose the difference $\T - \widehat{\T}$ and apply the above concentration results. In the decomposition, we need to bound terms of the form 
$\Exx^{-\frac{1}{2}} \Sxx^{\frac{1}{2}}  - \I$. Such bounds can be derived
from our assumption on $\norm{\Exx^{-\frac{1}{2}} \Sxx \Exx^{-\frac{1}{2}}
  - \I }$ using Lemma~\ref{lem:perturbation-matrix-square-root} below. 
This lemma is derived from the main result of~\citet{Mathias97a}, with
extra effort taken to understand the size of perturbation for which higher
order error terms can be safely ignored. 

\begin{lemma}[Perturbation of matrix square root] 
\label{lem:perturbation-matrix-square-root} 
Let $\HH \in \bbR^{d\times d}$ be positive definite, with eigenvalues
 in the range $[\sigma_{\min}, \sigma_{\max}]$ for some $\sigma_{\min}>0$. 
Let $\bTheta \in \bbR^{d\times d}$ be Hermitian, satisfying
$\norm{\HH^{-\frac{1}{2}} \bTheta \HH^{-\frac{1}{2}}}=1$.
Then for $\zeta \le \frac{3}{4} \sigma_{\max}^{-2} \sigma_{\min}^2$, we have 
\begin{gather*}
\norm{(\HH + \zeta \cdot \bTheta )^{\frac{1}{2}} \HH^{-\frac{1}{2}} - \I} \le C_d \cdot \zeta
\end{gather*}
where $C_d = \calO(\log d)$ is independent of $\zeta$. 
\end{lemma}

\begin{lemma}\label{lem:approx-error-erm}
Assume that we draw $N$ samples $\{(\x_i,\y_i)\}_{i=1}^N$ independently
from the underlying joint distribution $P(\x,\y)$ for computing the sample
covariance matrices in~\eqref{e:cov-empirical}, and $P(\x,\y)$ satisfies
Assumption~\ref{assump:bounded-covariances}
and the concentration property~\eqref{e:property-probability}. 
Then for $\nu \le \frac{1}{4} \gamma^2$, we have 
\begin{align*}
\abs{ \widehat{\rho}_1 - \rho_1} \le \norm{\T - \widehat{\T}} \le 4 C_d \cdot \nu
\end{align*}
where $C_d$ is the same constant in Lemma~\ref{lem:perturbation-matrix-square-root}. 
\end{lemma}

We note that the requirement of $\nu=\calO(\gamma^2)$ is not
too constraining, since the size of the perturbation $\nu$ is closely related to the statistical error, and we
are mainly interested in the regime of the statistical error going to
zero.
It is then straightforward to combine Lemma~\ref{lem:probability} and Lemma~\ref{lem:approx-error-erm}
to obtain the following sample complexities for the three distribution classes.
\begin{corollary}[Sample complexity for learning canonical
  correlation by ERM]
\label{cor:sample-complexity-canoncorr}
Let $\epsilon^\prime\in (0,1)$ and $\epsilon^\prime \le C_d \gamma^2$. Then for $N\ge N_0 \left(\frac{\epsilon^\prime}{4C_d}\right)$, i.e, 
\begin{align*}
& N \ge C \frac{d \log^2 d}{{\epsilon^\prime}^2}  \qquad\qquad\qquad   \text{for the sub-Gaussian class,} \\
& N \ge C \frac{d \log^{2 (1+r^{-1})} d}{{\epsilon^\prime}^{2(1+r^{-1})}}  \quad\qquad  \text{for the polynomial-tail class,} \\
& N \ge C \frac{\log^2 d}{{\epsilon^\prime}^2 \gamma^2}  \qquad\qquad\qquad\;\;  \text{for the bounded class,}
\end{align*}
we have with high probability that 
$\abs{ \widehat{\rho}_1 - \rho_1} \le \epsilon^\prime$.
\end{corollary}
\begin{remark}
Due to better concentration properties, the sample complexity for the sub-Gaussian and regular polynomial-tail classes are independent of the condition number $\frac{1}{\gamma}$ of the auto-covariances.
\end{remark}

\paragraph{Comparison to~\citet{Arora_17a}} 
In a parallel work by~\citet{Arora_17a}, the authors studied the top-k stochastic CCA for bounded inputs, and proposed
stochastic approximation-type algorithms with $\tilde{\calO} \left( \frac{1}{
    {\epsilon^\prime}^2 \gamma^2}  \right)$ sample-complexity upper bound 
for approximating the top canonical correlation.
We note, however, their stochastic algorithms are
derived from the convex relaxation of stochastic CCA, which lifts the
original problem into the space of matrices in $\bbR^{d_x \times d_y}$ and requires
a whitening operation (multiplying each fresh sample by 
$\Sxx^{-\frac{1}{2}}$ or $\Syy^{-\frac{1}{2}}$) and a projection operation
(onto the set of low 2-norm and nuclear-norm matrices) in each iteration, 
which are inefficient in high dimensions. 
Our work studies three different classes of input distributions in a
uniform manner\footnote{If we only study the case of bounded inputs, we
  can bypass Lemma~\ref{lem:perturbation-matrix-square-root} and the $\log
  d$ dependence in our bound can be reduced.}, 
with the goal of matching the statistical limits for the Gaussian inputs (see Section~\ref{sec:lower-bound}). The algorithms we
provide in the next sections require only elementary vector operations and
thus more practical for high dimensional data.

\subsection{Approximating the canonical directions}
\label{sec:erm-eigen-vector}

We now discuss the error in learning $(\uu^*,\vv^*)$ by ERM, when $\T$ has a singular value gap $\Delta > 0$. 
Let the nonzero singular values of $\T$ be $1 \ge \rho_1 \ge \rho_2 \ge \dots \ge \rho_r$, where $r = \text{rank}(\T) \le \min(d_x,d_y)$, and the corresponding (unit-length) singular vector pairs be $(\aa_1,\b_1),\dots,(\aa_r,\b_r)$. 
Define 
\begin{align} \label{e:C-def}
  \C = \left[
    \begin{array}{cc}
      \0 & \T \\
      \T^\top & \0 
    \end{array}
  \right] \in \bbR^{ d \times d }.
\end{align}

The eigenvalues of $\C$  are 
\begin{align*} 
\rho_1 \ge \dots \ge \rho_r > 0 = \dots = 0 > -\rho_r \ge \dots \ge -\rho_1,
\end{align*}
 with corresponding unit eigenvectors 
\begin{gather*}
   \frac{1}{\sqrt{2}}
   \left[ \begin{array}{c} \aa_1\\ \b_1 \end{array} \right], \ \dots,\ 
   \frac{1}{\sqrt{2}} 
   \left[ \begin{array}{c} \aa_r\\ \b_r \end{array} \right], \ \dots,\ 
   \frac{1}{\sqrt{2}} 
   \left[ \begin{array}{c} \aa_r\\ - \b_r \end{array} \right], \ \dots,\  
   \frac{1}{\sqrt{2}} 
   \left[ \begin{array}{c} \aa_1\\ - \b_1 \end{array} \right]. 
\end{gather*}
Thus, learning canonical directions $(\uu^*,\vv^*)$ reduces to learning the top eigenvector of $\C$.

We denote the empirical version of $\C$ by $\widehat{\C}$, and the singular vector pairs of $\widehat{\T}$ by $\{ (\widehat{\aa}_i, \widehat{\b}_i )\}$. Due to the block structure of $\C$ and $\widehat{\C}$, we have $\norm{\C - \widehat{\C}}=\norm{\T - \widehat{\T}}$. 
Let the ERM solution be $(\widehat{\uu}, \widehat{\vv})=\left( \Sxx^{-\frac{1}{2}} \widehat{\aa}_1, \Syy^{-\frac{1}{2}} \widehat{\b}_1 \right)$, which satisfy $\norm{\Sxx^{\frac{1}{2}} \widehat{\uu}} = \norm{\Syy^{\frac{1}{2}} \widehat{\vv}} = 1$. 
We now state the sample complexity for learning the canonical directions by ERM. 

\begin{theorem}\label{thm:approx-direction-erm}
Let $\epsilon \in (0,1)$ and $\epsilon \le \frac{16 C_d^2 \gamma^4}{\Delta^2}$. 
Then for $N\ge N_0 \left( \frac{\sqrt{\epsilon} \Delta}{16 C_d} \right)$, \ie,
\begin{align*}
&N \ge C \frac{d \log^2 d}{\epsilon \Delta^2}  \qquad\qquad\qquad\qquad   \text{for the sub-Gaussian class,} \\
&N \ge C \frac{d \log^{2 (1+r^{-1})} d}{\epsilon^{(1+r^{-1})} \Delta^2}  \qquad\qquad\quad  \text{for the regular polynomial-tail class,} \\
&N \ge C \frac{\log^2 d}{\epsilon \Delta^2 \gamma^2}  \qquad\qquad\qquad\qquad    \text{for the bounded class,}
\end{align*}
we have with high probability that 
$\text{align} \left( (\widehat{\uu},\widehat{\vv}); (\uu^*, \vv^*) \right) \ge 1 - \epsilon$.
\end{theorem}

\paragraph{Proof sketch\ } 
The proof of Theorem~\ref{thm:approx-direction-erm} consists of two steps.
We first bound the error between $\widehat{\C}$'s top eigenvector
$\frac{1}{\sqrt{2}} \left[ \begin{array}{c}\Sxx^{\frac{1}{2}}
                             \widehat{\uu}\\ \Syy^{\frac{1}{2}}
                             \widehat{\vv}\end{array}\right]$ and ${\C}$'s
                         top eigenvector $\frac{1}{\sqrt{2}}
                         \left[ \begin{array}{c}\Exx^{\frac{1}{2}} \uu^*
                                  \\ \Eyy^{\frac{1}{2}}
                                  \vv^* \end{array}\right]$ using a
                              standard result on perturbation of
                              eigenvectors, namely the Davis-Kahan $\sin
                              \theta$ theorem~\citep{DavisKahan70a} which
                              states $\sin^2 \theta \le
                              \frac{\norm{\C - \widehat{\C}}^2}{\Delta^2}
                              \le \frac{{\epsilon^\prime}^2}{\Delta^2}$
                              where $\theta$ is the angle between top
                              eigenvectors of $\C$ and $\widehat{\C}$. 
We then show that $\frac{1}{\sqrt{2}} \left[ \begin{array}{c}\Sxx^{\frac{1}{2}} \widehat{\uu}\\ \Syy^{\frac{1}{2}} \widehat{\vv}\end{array}\right]$ is very close to the ``correctly normalized'' $\frac{1}{\sqrt{2}} \left[ \begin{array}{c}\Exx^{\frac{1}{2}} \widehat{\uu} / \smallnorm{\Exx^{\frac{1}{2}} \widehat{\uu}} \\ \Eyy^{\frac{1}{2}} \widehat{\vv} / \smallnorm{\Eyy^{\frac{1}{2}} \widehat{\vv}} \end{array}\right]$, so the later still aligns well with the population solution. 

\paragraph{Comparison to prior analysis\ }
For the sub-Gaussian class, the tightest analysis of the sample complexity upper bound we are aware of was by~\citet{Gao_17a}. However, their proof relies on the assumption that $\rho_2 = o (\rho_1)$, \ie, they require that $\rho_2\ll \rho_1$. In contrast, we do not require this assumption, and our bound is sharp in terms of the gap $\Delta = \rho_1 - \rho_2$. 
Up to the $\log^2 d$ factor, our ERM sample complexity for the same loss matches the minimax lower bound $\frac{d}{\epsilon \Delta^2}$ given by ~\citet{Gao_17a} (see also Section~\ref{sec:lower-bound}). 
\section{Stochastic optimization for ERM}
\label{sec:stochastic-opt}

A disadvantage of the empirical risk minimization approach is that it can be time and memory consuming. To obtain the exact solution to~\eqref{e:cca-empirical}, we need to explicitly form and store the covariance matrices and to compute their singular value decompositions (SVDs); these steps have a time complexity of $\calO(N d^2 + d^3)$ and a memory complexity of $\calO(d^2)$. 

In this section, we study the stochastic optimization of the empirical objective, and show that the computational complexity is low: We just need to process a large enough dataset (with the same level of samples as ERM requires) nearly constant times in order to achieve small error with respect to the population objective. 
The basic algorithm we use here is the shift-and-invert meta-algorithm proposed by~\citet{Wang_16b}. 
However, in this section we provide refined analysis of the algorithm's time complexity than that provided by~\citet{Wang_16b}. We show that, using a better measure of progress and careful initializations for each least squares problem, the algorithm enjoys linear convergence (see Theorem~\ref{thm:total-time-stochastic-any-N}), \ie, the time complexity for achieving $\eta$-suboptimalilty in the empirical objective depends on $\log \frac{1}{\eta}$, whereas the result of~\citet{Wang_16b} has a dependence of $\log^2 \frac{1}{\eta}$.

We also note that the recent work of~\citet{Allen-ZhuLi16a} and~\citet{ZhuLi17a} have extended the ERM problem to extracting the top $k \ge 1$ pairs of canonical directions, and applied the technique of peeling/deflation together with shift-and-invert. However, their convergence rate for the fist pair of canonical directions does not improve that of~\citet{Wang_16b}.\footnote{See the second and third last lines of Table~1, and the last paragraph of Section~1.2 in~\citet{ZhuLi17a}: ``Our running time matches that of [29] when $k=1$''.}
As mentioned above, our result strictly improves that of~\citet{Wang_16b}, and in particular replaces the $\tilde{\calO} (\cdot)$ notation with the $\calO (\cdot)$ notation in total runtime, achieving true linear convergence.

\paragraph{Roadmap for this section} 
We first introduce the shift-and-invert power iterations and provide its iteration complexity, assuming that each matrix-vector multiplication or equivalently a convex least squares problem is solved to sufficient accuracy (Lemma~\ref{lem:progress-of-matrix-vector-multiplication}). We then show each least squares can be warm-started using rescaled estimates from the previous iteration (Lemma~\ref{lem:warm-start}). Finally, we plug in the time complexity of SVRG for each subproblem, and give runtime complexities for each distribution class which have different ``condition numbers'' (Corollary~\ref{cor:total-time-erm-stochastic}). 

The condition numbers depend on, among other things, the smallest eigenvalues of the covariance matrices, which are bounded away from zero as discussed below.   

\paragraph{Eigenvalues of empirical covariance}
According to the analysis of ERM from previous section, we have been working in the regime that the concentration parameter in~\eqref{e:property-probability} satisfies $\nu \le \frac{\gamma^2}{4} \le \frac{\gamma}{2}$. Thus in view of Assumption~\ref{assump:bounded-covariances}, we have with high probability that 
\begin{align*}
\norm{\Sxx - \Exx} = \norm{\Exx^{\frac{1}{2}} (\Exx^{-\frac{1}{2}} \Sxx \Exx^{-\frac{1}{2}} - \I) \Exx^{\frac{1}{2}}}
\le \norm{\Exx} \cdot \norm{\Exx^{-\frac{1}{2}}\Sxx\Exx^{-\frac{1}{2}} - \I} \le \frac{\gamma}{2}
\end{align*}
and similarly $\norm{\Syy - \Eyy} \le \frac{\gamma}{2}$. According to Weyl's inequality, these inequalities make sure eigenvalues of $\Sxx$ and $\Syy$ lie in $[\frac{\gamma}{2}, 1+\frac{\gamma}{2}]$, and consequently the involved subproblems are strongly-convex and can be solved efficiently. 

\subsection{Shift-and-invert power iterations} \label{s:alg-shift-and-invert}

Our algorithm runs the shift-and-invert power iterations on the following matrix
\begin{align}
  \label{e:M-def}
  \widehat{\M}_{\lambda} =
  \left(
    \lambda \I - \widehat{\C}
  \right)^{-1}
  =
  \left[ 
    \begin{array}{cc}
      \lambda \I & -\widehat{\T} \\
      -\widehat{\T}^\top & \lambda \I
    \end{array}
  \right]^{-1} 
\end{align}
where $\lambda>\widehat{\rho}_1$. 
It is straightforward to see that $\widehat{\M}_{\lambda}$ is positive definite with eigenvalues 
\begin{align*}
  \frac{1}{\lambda-\widehat{\rho}_1} \ge \dots \ge \frac{1}{\lambda-\widehat{\rho}_r} \ge \dots \ge \frac{1}{\lambda+\widehat{\rho}_r} \ge \dots \ge \frac{1}{\lambda+\widehat{\rho}_1},
\end{align*}
and has the same set of eigenvectors as $\widehat{\C}$.

Assume that there exists a singular value gap for $\widehat{\T}$ (this can be guaranteed by drawing sufficiently many samples so that the singular values of $\widehat{\T}$ are within a fraction of the gap $\Delta$ of ${\T}$), denoted as $\widehat{\Delta}=\widehat{\rho}_1  - \widehat{\rho}_2$. The key observation is that, as opposed to running power iterations on $\widehat{\C}$ (which is essentially done by~\citealt{Ge_16a}), $\widehat{\M}_{\lambda}$ has a large eigenvalue gap when $\lambda = \widehat{\rho}_1 + c (\widehat{\rho}_1  - \widehat{\rho}_2)$ with $c= \calO(1)$, and thus power iterations on $\widehat{\M}_{\lambda}$ converge more quickly. In particular, we assume for now the availability of an estimated eigenvalue $\lambda$ such that $\lambda - \widehat{\rho}_1 \in [l \widehat{\Delta}, u \widehat{\Delta}]$ where $0 < l <   u < 1$; locating such a $\lambda$ is discussed later in Remark~\ref{rmk:locate-lambda}.

Define 
\begin{align*}
\widehat{\A}_{\lambda} := \left[ \begin{array}{cc} \lambda \Sxx & - \Sxy \\ - \Sxy^\top & \lambda \Syy \end{array}\right], \qquad \quad
\widehat{\B} := \left[ \begin{array}{cc}  \Sxx & \0  \\ \0 & \Syy \end{array} \right],
\end{align*}
and we have $\widehat{\M}_{\lambda} = \widehat{\B}^{\frac{1}{2}} \widehat{\A}_{\lambda}^{-1} \widehat{\B}^{\frac{1}{2}}$. 
And by the relationship $\widehat{\A}_{\lambda}=\widehat{\B}^{\frac{1}{2}} \widehat{\M}_{\lambda}^{-1} \widehat{\B}^{\frac{1}{2}}$, eigenvalues of $\widehat{\A}_{\lambda}$ are bounded: 
\begin{gather*} 
 \sigma_{\max} \left( \widehat{\A}_{\lambda} \right) \le \sigma_{\max} \left(  \widehat{\M}_{\lambda}^{-1} \right) \cdot \sigma_{\max} \left(  \widehat{\B} \right) \le (\lambda+\widehat{\rho}_1) (1+\frac{\gamma}{2}), \\
 \sigma_{\min} \left( \widehat{\A}_{\lambda} \right) \ge \sigma_{\min} \left(  \widehat{\M}_{\lambda}^{-1} \right) \cdot \sigma_{\min} \left(  \widehat{\B} \right) 
\ge (\lambda-\widehat{\rho}_1) \gamma / 2.
\end{gather*}
It is convenient to study the convergence in the concatenated variables
 \begin{align*}
  \w_{t}  := \frac{1}{\sqrt{2}} \left[ \begin{array}{c} \uu_{t}  \\ \vv_{t} \end{array}\right], \quad 
  \rr_{t}& :=  \widehat{\B}^{\frac{1}{2}} \w_{t} = \frac{1}{\sqrt{2}} \left[ \begin{array}{c} \Sxx^{\frac{1}{2}} \uu_{t}  \\ \Syy^{\frac{1}{2}} \vv_{t} \end{array}\right].
\end{align*}
Define the following quantities using the ERM solution 
\begin{align*}
  \widehat{\w}  := \frac{1}{\sqrt{2}} \left[ \begin{array}{c} \widehat{\uu}  \\ \widehat{\vv} \end{array}\right], \qquad 
  \widehat{\rr} :=  \widehat{\B}^{\frac{1}{2}} \widehat{\w} = \frac{1}{\sqrt{2}} \left[ \begin{array}{c} \Sxx^{\frac{1}{2}} \widehat{\uu}  \\ \Syy^{\frac{1}{2}} \widehat{\vv} \end{array}\right],
\end{align*}
which satisfy ${\widehat{\w}}^\top \widehat{\B} \widehat{\w} = 1$ and  ${\widehat{\rr}}^\top \widehat{\rr} = 1$ respectively.

\subsection{Convergence of inexact shift-and-invert}

Our algorithm iteratively applies the approximate matrix-vector multiplications: for $t=0,1,\dots$
\begin{align} \label{e:power-iterations-uv}
  \rr_{t+1} \approx \widehat{\M}_{\lambda} \rr_{t}, 
  \qquad \Longleftrightarrow \qquad
  \w_{t+1} \approx \widehat{\A}_{\lambda}^{-1} \widehat{\B} \w_{t}. 
\end{align}
This equivalence allows us to directly work with $(\uu_{t},\vv_{t})$ and avoids computing $\Sxx^{\frac{1}{2}}$ or $\Syy^{\frac{1}{2}}$ explicitly. Note that we do not perform normalizations of the form $\w_t \leftarrow \w_t / \norm{\widehat{\B}^{\frac{1}{2}} \w_t}$ at each iteration as done by~\citet{Wang_16b} (Phase-I of their SI meta-algorithm); the length of each iterate is irrelevant for the purpose of optimizing the alignment between vectors and we could always perform the normalization in the end to satisfy the length constants. Exact power iterations is known to converge linearly when there exist an eigenvalue gap~\citep{GolubLoan96a}.

The matrix-vector multiplication $ \widehat{\A}_{\lambda}^{-1} \widehat{\B} \w_{t}$ is equivalent to solving the least squares problem 
\begin{align} \label{e:lsq}
  \min_{\w}\; f_{t+1} (\w) := \frac{1}{2} \w^\top \widehat{\A}_{\lambda} \w - \w^\top \widehat{\B} \w_{t}
\end{align}
whose unique solution is  $\w_{t+1}^* = \widehat{\A}_{\lambda}^{-1} \widehat{\B} \w_{t}$ with the optimal objective $f_{t+1}^* = - \frac{1}{2} \w_{t}^\top \widehat{\B} \widehat{\A}_{\lambda}^{-1} \widehat{\B} \w_{t}$. 
Of course, solving the problem exactly is costly and we will apply stochastic gradient methods to it. We will show that, when the least squares problems are solved accurately enough, the iterates are of the same quality as those of the exact solutions and  enjoys linear convergence. 

We begin by introducing the measure of progress for the iterates. Denote the eigenvalues of $\widehat{\M}_{\lambda}$ by $\beta_1\ge \beta_2\ge \dots \ge \beta_d$, with corresponding eigenvectors $\p_1,\dots,\p_d$ forming an orthonormal basis of $\bbR^d$. 
Recall that $\p_1=\widehat{\rr}$, $\p_i^\top \widehat{\M}_{\lambda} \p_i=\beta_i$ for $i=1,\dots,d$, and $\p_i^\top \widehat{\M}_{\lambda} \p_j=0$ for $i\neq j$. 

We therefore can write each iterate as a linear combination of the eigenvectors:
$\frac{\rr_t}{\norm{\rr_t}} = \sum_{i=1}^d \xi_{ti} \p_i$, where $\xi_{ti}= \frac{\rr_t^\top \p_i}{\norm{\rr_t}}$ for $i=1,\dots,d$, and $\sum_{i=1}^d \xi_{ti}^2 = 1$. 
The potential function we use to evaluate the progress of each iteration is 
\begin{align*}
  G (\rr_{t})  = \frac{\normhatM{\PP_{\perp} \frac{\rr_{t}}{\norm{\rr_{t}}}}}{\normhatM{\PP_{\parallel} \frac{\rr_{t}}{\norm{\rr_{t}}}}} 
  = \frac{ \sqrt{ \sum_{i=2}^d  \xi_{ti}^2 / \beta_i }}{ \sqrt{ \xi_{t1}^2 / \beta_1}}, 
\end{align*}
where $\PP_{\perp}$ and $\PP_{\parallel}$ denote projections onto the subspaces perpendicular and parallel to $\widehat{\rr}$ respectively. 

The same potential function was used by~\citet{Garber_16a} for analyzing the convergence of shift-and-invert for PCA. The potential function is invariant to the length of $\rr_{t}$, and is equivalent to the criterion $\abs{\tan \theta_t} := \frac{\sqrt{\sum_{i=2}^d  \xi_{ti}^2}}{\sqrt{ \xi_{t1}^2}}$ where $\theta_t$ is the angle between $\rr_{t}$ and $\widehat{\rr}$: in the following sense:
\begin{align*} 
  \abs{\sin \theta_t} = \sqrt{\sum_{i=2}^d  \xi_{ti}^2} 
  \le \sqrt{\frac{\beta_1}{\beta_2}} \abs{\tan \theta_t} \le  G (\rr_{t}) \le \sqrt{\frac{\beta_1}{\beta_d}} \abs{\tan \theta_t} . 
\end{align*}

The lemma below shows that under the iterative scheme~\eqref{e:power-iterations-uv}, $\{G(\rr_t)\}_{t=1,\dots}$ converges linearly to $0$.

\begin{lemma} \label{lem:progress-of-matrix-vector-multiplication}
  Let $\eta \in (0,1)$. Assume that for each approximate matrix-vector multiplication, we solve the least squares problem so accurately that the approximate solution $\w_{t+1}$ satisfies
  \begin{gather} \label{e:epsilon-t}
    \epsilon_t := \frac{f_{t+1} (\w_{t+1}) - f_{t+1}^*}{\w_t^\top \widehat{\B} \w_t} 
    \le  \min \left( {\sum_{i=2}^d \xi_{ti}^2 / \beta_i},\ {\xi_{t1}^2 / \beta_1} \right) \cdot  \frac{\left( \beta_1 - \beta_2 \right)^2}{32}.
  \end{gather}
  Let $T = \ceil{ \log_{\frac{7}{5}} \left( \frac{G(\rr_0)}{\eta} \right) }$. Then we have $\abs{\sin \theta_t} \le G(\rr_t) \le \eta$ for all $t\ge T$.
\end{lemma}

\subsection{Bounding initial error for least squares}

It is natural to use an initialization of the form $\alpha \w_{t}$ for minimizing $f_{t+1}(\w)$. The following lemma provides the optimal $\alpha$ and the resulting initial suboptimality, see detailed analysis in Appendix~\ref{sec:erm-least-squares-initial-error}.
\begin{lemma}[Warm start for least squares]
  \label{lem:warm-start}
  Initializing $\min_{\w}\, f_{t+1} (\w)$ with $\alpha_t^* \w_{t}$ where $\alpha_{t}^* = \frac{\w_{t}^\top \widehat{\B} \w_{t}}{\w_{t}^\top \widehat{\A}_{\lambda} \w_{t}}$, it suffices to set the ratio between the initial and the final error to be 
  $  64 \cdot \max \left( 1, G (\rr_{t}) \right) $
so that~\eqref{e:epsilon-t} is satisfied.
\end{lemma}
This result indicates that in the converging stage ($G(\rr_{t})\le 1$), we just need to set the ratio between the initial and the final error to the constant 64 (and set it to be the constant $64 G (\rr_0)$ before that). 
This will ensure that the time complexity of least squares has no dependence on the final error $\epsilon$.

\subsection{Solving the least squares by SGD}

The least squares objective~\eqref{e:lsq} is the sum of $N$ functions: $f_{t+1} (\w) = \frac{1}{N} \sum_{i=1}^N f_{t+1}^i(\w)$ where
\begin{align} \label{e:lsq-finite-sum}
  f_{t+1}^i (\w) =
  \frac{1}{2}
  \w^\top 
  \left[
    \begin{array}{cc}
      \lambda  \x_i \x_i^\top  & - \x_i \y_i^\top \\
      - \y_i \x_i^\top & \lambda \y_i \y_i^\top
    \end{array}
  \right]
  \w   - \w^\top 
  \left[
    \begin{array}{cc}
      \Sxx  & \0 \\
      \0 & \Syy
    \end{array}
  \right] \w_{t} .
\end{align}
There has been much recent progress on developping linearly convergent stochastic algorithms for solving finite-sum problems. We use SVRG~\citep{JohnsonZhang13a} here due to its algorithmic simplicity and memory efficiency; in the next section, we will be using the ``online'' version of SVRG for stochastic CCA in the streaming setting. Note that although $f_{t+1} (\w)$ is convex, each component $f_{t+1}^i$ may not be convex. 

We provide the time complexity of SVRG for this case (based on~\citealp[Appendix~B]{GarberHazan15c}), as well as the ``condition number'' for the three classes of distributions in Appendix~\ref{sec:lsq-finite-sum} and~\ref{sec:bounding-kappa-erm} respectively.

\subsection{Total time complexity}

We first provide the runtime for solving the empirical objective using the (offline) shift-and-invert CCA algorithm.
\begin{theorem} \label{thm:total-time-stochastic-any-N}
  Let $\eta \in (0,1)$. Draw $N$ samples for ERM such that $\sigma_{\min}(\Sxx) \ge \frac{\gamma}{2}$ and  $\sigma_{\min}(\Syy) ) \ge \frac{\gamma}{2}$. Initialize $\w_0 = \frac{\tw_0}{\sqrt{\tw_0^\top \widehat{\B} \tw_0}}$ where entries of $\tw_0 \in \bbR^d$ are randomly sampled from the standard Gaussian distribution. 
Then with high probability, offline shift-and-invert outputs an $(\uu_T,\vv_T)$ satisfying $\min\left( \frac{ {\uu_T}^\top \Sxx \widehat{\uu}}{\smallnorm{\Sxx^{\frac{1}{2}} {\uu_T}}},\, \frac{ {\vv_T}^\top \Syy \widehat{\vv}}{\smallnorm{\Syy^{\frac{1}{2}} {\vv_T}}} \right) \ge 1-\eta$ in total time
  \begin{align*}
    & \calO \left( d \left( N + \frac{d^2}{\widehat{\Delta}^2 \gamma^2} \right) \log \frac{d}{\widehat{\Delta} \gamma} \log \frac{d}{\widehat{\Delta} \gamma \eta} \right) \qquad\quad \text{for sub-Gaussian/polynomial-tail,} \\
    & \calO \left( d \left( N + \frac{1}{\widehat{\Delta}^2 \gamma^2} \right) \log \frac{d}{\widehat{\Delta} \gamma} \log \frac{d}{\widehat{\Delta} \gamma \eta} \right)  \qquad\quad \text{for the bounded class.} 
  \end{align*}
\end{theorem}
We have already shown in Theorem~\ref{thm:approx-direction-erm} that the ERM solution aligns well with the population solution. By drawing slighly more samples and requiring our algorithm to find an approximate solution that aligns well with the ERM solution, we can guarantee high alignment for the approximate solution as shown in the following corollary. 

\begin{corollary} \label{cor:total-time-erm-stochastic}
  Let $\epsilon \in (0,1)$ and $\epsilon \le \frac{64 C_d^2 \gamma^4}{\Delta^2}$. Draw $N=N_0 \left(\frac{\sqrt{\epsilon} \Delta}{32 C_d} \right)$ samples for the ERM objective, and use the initialization strategy in Theorem~\ref{thm:total-time-stochastic-any-N}. 
Then with high probability, the total time for offline shift-and-invert to output $(\uu_T,\vv_T)$ with $\text{align} \left( (\uu_T, \vv_T); (\uu^*, \vv^*) \right) \ge 1-\epsilon$ is 
  \begin{align*}
    & \calO \left( d \left( \frac{d \log^2 d}{\epsilon \Delta^2} + \frac{d^2}{\Delta^2 \gamma^2} \right) \log \frac{d}{\Delta \gamma} \log \frac{d}{\Delta \gamma \epsilon} \right)  \qquad\qquad\qquad\! \text{for  sub-Gaussian,} \\
    & \calO \left( d \left( \frac{d \log^{2 (1+r^{-1})} d}{\epsilon^{(1+r^{-1})} \Delta^2}  + \frac{d^2}{\Delta^2 \gamma^2} \right) \log \frac{d}{\Delta \gamma} \log \frac{d}{\Delta \gamma \epsilon} \right) \qquad\;\;  \text{for polynomial-tail,} \\
    & \calO \left( d \left( \frac{\log^2 d}{\epsilon \Delta^2 \gamma^2} + \frac{1}{\Delta^2 \gamma^2} \right) \log \frac{d}{\Delta \gamma} \log \frac{d}{\Delta \gamma \epsilon} \right) \qquad\qquad\qquad     \text{for the bounded class}.
  \end{align*}
\end{corollary}

The $\epsilon$-dependent term is near-linear in the ERM sample complexity $N (\epsilon,\Delta,\gamma)$ and is also the dominant term in the total runtime (when $\epsilon = o (\gamma^2)$ for the first two classes). 
For sub-Gaussian/regular polynomial-tail classes, we incur an undesirable $d^2$ dependence for the least squares problem's condition number 
(see more details in Appendix~\ref{sec:lsq-finite-sum}), mainly due to weak concentration regarding the data norm (we have stronger concentration for the streaming setting discussed next). One can alleviate the issue of large condition number using accelerated SVRG~\citep{Lin_15a}. 

\begin{remark} \label{rmk:locate-lambda}
  We have assumed so far the availability of $\lambda = \widehat{\rho}_1 + c (\widehat{\rho}_1  - \widehat{\rho}_2)$ with $c=\calO(1)$ for shift-and-invert to work. There exists an efficient algorithm for locating such an $\lambda$, see the \textbf{repeat-until} loop of Algorithm~3 in~\citet{Wang_16b}. This procedure computes $\calO \left( \log  \frac{1}{\Delta} \right)$ approximate matrix-vector multiplications, and its time complexity does not depend on $\epsilon$ as we only want to achieve good estimate of the top eigenvalue (and not the top eigenvector). So the cost of locating $\lambda$ is not dominant in the total runtime. 
\end{remark}

\section{Streaming shift-and-invert CCA}
\label{sec:alg-shift-and-invert-online}

A disadvantage of the ERM approach is that we need to store all the samples in order to go through the dataset multiple times. We now study the shift-and-invert algorithms in the streaming setting in which we draw samples from the underlying distribution $P(\x,\y)$ and process them once. Clearly, the streaming approach requires only $\calO(d)$ memory. 

In this section, we assume the availability of a $\lambda = \rho_1 + c
\Delta$, where  $0 < c < 1$.\footnote{Based on the same intuition given in
Remark~\ref{rmk:locate-lambda}, we believe that a procedure similar to
that of~\citet{Wang_16b} also works in the streaming setting and the cost
in locating $\lambda$ is not dominant, although we do not have a formal
analysis.}
Our algorithm is the same as in the ERM case, except that we now directly work with the population covariances through fresh samples instead of their empirical estimates. With slight abuse of notation, we use $(\A_{\lambda}, \B, \M_{\lambda})$ to denote the population version of $(\widehat{\A}_{\lambda}, \widehat{\B}, \widehat{\M}_{\lambda})$:
\begin{gather*}
  \A_{\lambda} := \left[ \begin{array}{cc} \lambda \Exx & - \Exy \\ - \Exy^\top & \lambda \Eyy \end{array}\right], \qquad 
  \B := \left[ \begin{array}{cc}  \Exx &  \0 \\ \0 & \Eyy \end{array}
  \right],  \qquad  
  \M_{\lambda} = \B^{\frac{1}{2}} \A_{\lambda}^{-1} \B^{\frac{1}{2}},
\end{gather*}
use $\{ \left(\beta_i, \p_i\right) \}_{i=1}^d$ to denote the eigensystem of $\M_{\lambda}$, and use $(\uu_t, \vv_t)$ as well as 
\begin{align*}
  \w_{t} =\frac{1}{\sqrt{2}} \left[\begin{array}{c} \uu_t \\ \vv_t \end{array}\right],  \quad
  \rr_{t} = \B^{\frac{1}{2}} \w_{t} = \frac{1}{\sqrt{2}} \left[\begin{array}{c} \Exx^{\frac{1}{2}} \uu_t \\ \Eyy^{\frac{1}{2}} \vv_t \end{array}\right], 
\end{align*}
$t=0,\dots$ 
to denote the iterates of our algorithm. Also, define $\xi_{ti}$, $\theta_t$ and $G(\rr_{t})$ similarly as in Section~\ref{sec:stochastic-opt}.

\paragraph{Handling normalizations} 
It is sufficient to achieve high alignment between 
\begin{align*}
  \frac{\rr_{T}}{\norm{\rr_{T}}} = \left[\begin{array}{c} \Exx^{\frac{1}{2}} \uu_T  \\ \Eyy^{\frac{1}{2}} \vv_T  \end{array}\right] \big/ \sqrt{{\uu}_T^\top \Exx {\uu}_T + {\vv}_T ^\top \Eyy {\vv}_T} 
\end{align*}
where $(\uu,\vv)$ are normalized jointly, 
and $\rr^* = \frac{1}{\sqrt{2}} \left[\begin{array}{c} \Exx^{\frac{1}{2}} \uu^* \\ \Eyy^{\frac{1}{2}} \vv^* \end{array}\right]$
where $(\uu,\vv)$ are normalized separately.
According to the lemma below, this would imply high alignment between $\frac{1}{\sqrt{2}} \left[\begin{array}{c} \Exx^{\frac{1}{2}} \uu_T / \smallnorm{\Exx^{\frac{1}{2}} \uu_T} \\ \Eyy^{\frac{1}{2}} \vv_T / \smallnorm{\Eyy^{\frac{1}{2}} \vv_T} \end{array}\right]$ and $\rr^*$ which is our final goal.

\begin{lemma}[Conversion from joint alignment to separate alignment] \label{lem:online-transfer-align} Let $\eta\in (0,1)$. If the output $(\uu_T, \vv_T)$ of our online shift-and-invert algorithm satisfy that
\begin{align*} 
 \frac{1}{\sqrt{2}} \cdot 
  \frac{{ {\uu^*}^\top \Exx \uu_T + {\vv^*}^\top \Eyy \vv_T }}
  {\sqrt{{\uu}_T^\top \Exx {\uu}_T + {\vv}_T ^\top \Eyy {\vv}_T}}
  \ge 1 - \frac{\eta}{4} , 
\end{align*}
  we also have 
\begin{align*}
 \text{align} \left( (\uu_T, \vv_T); (\uu^*, \vv^*) \right) = \frac{1}{2} \left( \frac{{\uu^*}^\top \Exx \uu_T}{\sqrt{\uu_T^\top \Exx \uu_T}} + \frac{{\vv^*}^\top \Eyy \vv_T}{\sqrt{\vv_T^\top \Eyy \vv_T}} \right) \ge 1 - \eta .
\end{align*}
\end{lemma}
Note that Lemma~\ref{lem:online-transfer-align} improves over a similar result by~\citet[Theorem~5]{Wang_16b}, which requires the joint alignment to be $\calO(\eta^2)$-suboptimal for the separate alignment to be $\calO(\eta)$-suboptimal.

\subsection{Solving least squares by streaming SVRG}

Turning to the streaming algorithm, the least squares problem at iteration $t+1$, 
is now a stochastic program:
\begin{align*} 
  \min_{\w}\; f_{t+1} (\w) = \frac{1}{2} \w^\top {\A}_{\lambda} \w - \w^\top {\B} \w_{t} = \bbE \left[ \phi_{t+1} (\w; \x, \y) \right] 
\end{align*}
where 
$  \phi_{t+1} (\w; \x, \y):= 
\frac{1}{2}
\w^\top 
\left[
  \begin{array}{cc}
    \lambda  \x \x^\top  & - \x \y^\top \\
    - \y \x^\top & \lambda \y \y^\top
  \end{array}
\right]
\w  - \w^\top 
\left[
  \begin{array}{cc}
    \x \x^\top  & \0 \\
    \0 & \y \y^\top
  \end{array}
\right] \w_{t} $,
and the expectation is computed over $P(\x,\y)$. The optimal solution to this stochastic program is $\w_{t+1}^* = \A_{\lambda}^{-1} \B \w_t $. 

\begin{algorithm}[t]
  \caption{Streaming SVRG for $\min_{\w}\ f(\w)$.}
  \label{alg:streaming-svrg}
  \renewcommand{\algorithmicrequire}{\textbf{Input:}}
  \renewcommand{\algorithmicensure}{\textbf{Output:}}
  \begin{algorithmic}
    \REQUIRE Initialization $\w^0 = \0$, stepsize scaling factor
    $s=\frac{1}{352}$, $(\mu, S, \sigma^2)$ are respectively the strong
    convexity, streaming smoothness, and streaming variance given in
    Lemma~\ref{lem:streaming-parameters}. 
    \FOR{$\tau=1,\dots,\Gamma$}
    \STATE $\bar{\z} \leftarrow \w^{\tau-1}$
    \STATE $m_{\tau} \leftarrow \ceil{\frac{44^2 S}{\mu}}, \quad k_{\tau} \leftarrow \max\left( \ceil{\frac{44 S}{\mu }}, \, \ceil{\frac{20 \sigma^2 \cdot 2^{\tau-1}}{\beta_1 \norm{\rr_{t}}^2}} \right)$
    \STATE Draw $k_{\tau}$ samples $(\x_1, \y_1),\dots,(\x_{k_{\tau}}, \y_{k_{\tau}})$ and estimate the batch gradient
    \begin{align*}
      \g \leftarrow \frac{1}{k_{\tau}} \sum\nolimits_{i=1}^{k_{\tau}} \nabla \phi (\bar{\z}; \x_i,\y_i)
    \end{align*}
    Sample $\widetilde{m}_{\tau}$ uniformly at random from $\{1,\dots, {m}_{\tau}\}$
    \STATE $\z\leftarrow \bar{\z}$
    \FOR {$i=1,\dots, \widetilde{m}_{\tau}$}
    \STATE Draw sample  $(\x_i, \y_i)$
    \STATE $\z \leftarrow \z - \frac{s}{S} \left( \nabla \phi (\z; \x_i,\y_i) 
      - \nabla \phi (\bar{\z}; \x_i,\y_i) + \g \right)$ 
    \ENDFOR
    \STATE $\w^\tau \leftarrow \z$
    \ENDFOR
    \ENSURE Return $\w^{\Gamma}$ as the approximate solution.
  \end{algorithmic}
\end{algorithm}

Due to the high sample complexity of accurately estimating $\alpha_t^* = \frac{\w_{t}^\top {\B} \w_{t}}{\w_{t}^\top {\A}_{\lambda} \w_{t}}$ in the streaming setting, we instead initialize each linear systems with the zero vector. With this initialization, we have
\begin{align} \label{e:streaming-initialization}
  f_{t+1} (\0) - f_{t+1}^* & = 0 - \left( - \frac{1}{2} \w_{t}^\top {\B} {\A}_{\lambda}^{-1} {\B} \w_{t} \right)  = \frac{\rr_{t}^\top {\M}_{\lambda} \rr_{t}}{2} \le \frac{\beta_1 \norm{\rr_{t}}^2}{2}.
\end{align}
We then solve the linear system with the streaming SVRG algorithm proposed
by~\citet{Frostig_15b}, as detailed in Algorithm~\ref{alg:streaming-svrg}.
This is the same approach taken by~\citet{Garber_16a} for streaming PCA, and our analysis follows the same structure. 
Streaming SVRG is a natural choice here since it is the ``online''
version of the SVRG algorithm for optimizing empirical objectives and
enjoys the same algorithmic simplicity and low computational complexity.
Moreover, for stochastic least squares problems, streaming SVRG is shown
to have the same sample complexity as solving the ERM problem~\citep{Frostig_15b}, which
aligns well with our goal of an overall sample efficient algorithm. 
With this choice, the final algorithm is very similar to the stochastic
optimization algorithm in Section~\ref{sec:stochastic-opt}, except that
fresh samples are used for each update.  

To analyze the sample complexity of streaming SVRG, we first calculate the streaming smoothness and streaming variance parameters for the three classes of distributions. 
\begin{lemma}[Parameters of streaming SVRG] \label{lem:streaming-parameters}
  For any $\w,\w^\prime \in \bbR^d$, we have
  \begin{itemize}
  \item Strong convexity:
    \begin{align*}
    \hspace*{-3em}  f_{t+1} (\w) \ge f_{t+1} (\w^\prime) + \left<\nabla f_{t+1} (\w^\prime), \w - \w^\prime \right> + \frac{\mu}{2} \norm{\w - \w^\prime}^2,
    \end{align*}
  \item Streaming smoothness:
    \begin{align*}
     \hspace*{-3em} \bbE \left[ 
        \norm{ \nabla \phi_{t+1} (\w) - \nabla \phi_{t+1} (\w_{t+1}^*) }^2 \right] 
      \le 2 S  \left( f_{t+1} (\w) - f_{t+1}^* \right),
    \end{align*}
  \item Streaming variance:
    \begin{align*}
     \hspace*{-3em} \bbE \left[ \frac{1}{2} \norm{\nabla \phi (\w_{t+1}^*)}^2_{\left( \nabla^2 f(\w_{t+1}^*)\right)^{-1}} \right] \le \sigma^2.
    \end{align*}
\end{itemize}
    Here $\mu:= \frac{\gamma}{\beta_1} \ge C \Delta \gamma$ for some $C>0$, and 
    \begin{align*}
S = \calO \left( \frac{d \beta_1}{\gamma} \right), \quad \sigma^2 = \calO \left( d \beta_1^3 \norm{\rr_{t}}^2 \right)
\end{align*}
for sub-Gaussian/regular polynomial-tail classes, and 
\begin{align*}
S = \calO \left( \frac{\beta_1}{\gamma} \right), \quad \sigma^2= \calO \left( \frac{\beta_1^3 \norm{\rr_{t}}^2}{\gamma^2} \right) 
\end{align*} 
for the bounded class.
  \end{lemma}

  The proof of Lemma~\ref{lem:streaming-parameters} is somewhat technical
  for the sub-Gaussian/regular polynomial-tail classes, which repeatedly
  applies the concentration properties of these two classes. But this
  lemma is the key for the sample complexity of our streaming algorithm to
  match the lower bound in the case of sub-Gaussian inputs: since we
  always draw fresh samples in the streaming setting, the ``condition
  number'' $S/\mu$ for these two classes depend on $d$ only linearly (as
  opposed to quadratically in approximate ERM). 
These quantities determine the number of samples to be used in each round
$\tau$ of Algorithm~\ref{alg:streaming-svrg}:
$m_{\tau}$ is on the order of the condition number, and with $m_{\tau}$
stochastic updates, one can reduce the suboptimality by a constant factor in each round;
$\kappa_{\tau}$ has to eventually increase geometrically to make sure the
variance is reduced at the same pace.

  Based on these quantities, we can apply the structural result of~\citet{Frostig_15b} and give the sampling complexity for driving the final suboptimality to $\eta_t$ times the initial suboptimality in~\eqref{e:streaming-initialization}.

  \begin{lemma}[Sample complexity of streaming SVRG for least squares] 
    \label{lem:streaming-svrg-sample-complexity-for-f}
    Let $\eta_t \in (0,1)$. Applying streaming-SVRG in Algorithm~\ref{alg:streaming-svrg} to $\min_{\w} \ f_{t+1} (\w)$ with initialization $\0$, we have 
\begin{align*}
\bbE \left[ f_{t+1} (\w^{\tau}) - f_{t+1}^* \right] \le \eta_t \left( \frac{\beta_1 \norm{\rr_{t}}^2}{2} \right) 
\end{align*}
    for $\tau\ge \Gamma = \calO \left( \log \frac{1}{\eta_t} \right)$. 
    The sample complexity of the first $\Gamma$ iterations is $\calO \left( \frac{d}{\Delta^2 \eta_t} + \frac{d}{\Delta^2 \gamma^2} \log \frac{1}{\eta_t} \right)$ 
    for the sub-Gaussian/regular polynomial-tail classes, and $\calO \left(  \frac{1}{\Delta^2 \gamma^2 \eta_t}  \right)$ for the bounded class.
  \end{lemma}

  Based on the linear convergence of shift-and-invert, we need only solve
  $\calO \left( \log \frac{1}{\epsilon} \right)$ linear systems, and we
  can bound $\frac{1}{\eta}$ by a geometrically increasing series where
  the last term is $\calO \left( \frac{1}{\epsilon} \right)$ (so the sum
  of this truncated series is still $\calO \left( \frac{1}{\epsilon} \right)$). This results in the following total sample complexity.

  \begin{theorem}[Total sample complexity of streaming shift-and-invert CCA]
    \label{thm:sample-complexity-for-online-cca}
    Let $\epsilon\in (0,1)$. After solving $T = \calO \left( \log \frac{1}{\epsilon} \right)$ linear systems to sufficient accuracy, streaming shift-and-invert CCA algorithm outputs $(\uu_T,\vv_T)$ with $\text{align} \left( (\uu_T, \vv_T); (\uu^*, \vv^*) \right) \ge 1-\epsilon$. Our algorithm processes each sample in $\calO(d)$ time, and has a total sample complexity of 
    \begin{align*}
      & \calO \left( \frac{d}{\epsilon  \Delta^2} + \frac{d}{\Delta^2 \gamma^2} \log^2 \frac{1}{\epsilon} \right) \qquad\quad \text{for the sub-Gaussian/regular polynomial-tail classes,} \\
      & \calO \left( \frac{1}{\epsilon  \Delta^2 \gamma^2} \right) \qquad\qquad\qquad\qquad\; \text{for the bounded class.}
    \end{align*}
  \end{theorem}

  Interestingly, the sample complexity of our streaming CCA algorithm (assuming the parameter $\lambda$) improves over that of ERM we showed in Theorem~\ref{thm:approx-direction-erm}: it removes small $\log d$ factors for all classes, and most remarkably achieves polynomial improvement in $\epsilon$ for the regular polynomial-tail class. 
  This is due to the fact that the sample complexity of streaming SVRG basically only uses the moments, and does not require concentration of the whole covariance in Lemma~\ref{lem:probability}. 
  As a result, it is not clear if our analysis of ERM is the tightest possible. 

\subsection{Lower bound for Gaussian inputs}
\label{sec:lower-bound}

Consider the following Gaussian distribution named \emph{single canonical pair model}~\citep{Chen_13e}:
\begin{align} \label{e:lower-bound-cov}
\left[\begin{array}{c}\x \\ \y\end{array}\right] \sim 
\calN \left(\0, \, \left[
\begin{array}{cc}
\I & \Delta  \bphi \bpsi^\top  \\
\Delta \bpsi \bphi^\top &  \I 
\end{array} 
\right]  \right),
\end{align}
where $\norm{\bphi}=\norm{\bpsi} = 1$. 
It is straightforward to check that $\T = \Exy = \Delta \bphi \bpsi^\top$ for such a distribution. Observe that $\T$ is of rank one and has a singular value gap $\Delta$, and the single pair of canonical directions are $\left( \uu^*, \vv^* \right) = \left(  \bphi,\, \bpsi \right)$. Denote this class of model by $\calF (d_x, d_y, \Delta)$. 
We have the following minimax lower bound for CCA under this model, which is an application of the result of~\citet{Gao_17a} for sparse CCA (by using rank $r=1$ and hard sparsity, \ie, $q=0$ and sparsity level $d$ in their Theorem~3.2). 

\begin{lemma}[Lower bound for single canonical pair model]
Suppose the data is generated by the single canonical pair model. Let $(\uu,\, \vv)$ be some estimate of the canonical directions $(\uu^*,\, \vv^*)$ based on $N$ samples. Then, there is a universal constant $C$, so that for $N$ sufficiently large, we have:
\begin{align*}
\inf_{\uu,\vv}\; \sup_{ \stackrel{\uu^*,\vv^* \in }{\calF (d_x, d_y, \Delta)}} \; \bbE \left[ 1 - \text{align} \left( (\uu_T, \vv_T); (\uu^*, \vv^*) \right) \right] \ge C \frac{d}{\Delta^2 N}.
\end{align*}
\end{lemma}

This lemma implies that, to estimate the canonical directions up to $\epsilon$-suboptimality in our measure of alignment, we expect to use at least $\calO \left( \frac{d}{\epsilon \Delta^2} \right)$ samples. 
We therefore observe that, for Gaussian inputs, the sample complexity of
the our streaming algorithm matches that of the minimax rate of CCA, up to
small factors.

In Table~\ref{t:summary}, we collect the complexities of different
approaches, namely exact optimization of ERM (Section~\ref{sec:erm}),
stochastic optimization of ERM with shift-and-invert
(Section~\ref{sec:stochastic-opt}), and streaming shift-and-invert
(Section~\ref{sec:alg-shift-and-invert-online}). We observe that while all
three approaches are sample efficient (up to small factors), stochastic
and streaming algorithms are more efficient in time and memory. 

\begin{table}[t]
\centering
\caption{Summary of sample, time (measured in floating point operations), and memory complexities of different
  approaches, in terms of $(d,\,\Delta,\,\epsilon)$, for stochastic CCA
  with Gaussian inputs. 
We give the dominant term in complexities as $\epsilon \rightarrow 0$.
Note that the time complexity of exact ERM is dominated by forming the eigen-system,
while the memory complexity of ERM is dominated by saving the dataset.}
\label{t:summary}
\begin{tabular}{@{}|c|c|c|c|@{}}
\hline
Method & Sample & Time & Memory \\
\hline
Exact ERM & $\tilde{\calO} \rbr{\frac{d}{\epsilon \Delta^2}}$ 
&  $\tilde{\calO} \rbr{\frac{d^3}{\epsilon \Delta^2}}$ 
&  $\tilde{\calO} \rbr{\frac{d^2}{\epsilon \Delta^2}}$ \\ \hline
\caja{c}{c}{Approximate ERM\\by shift-and-invert} & $\tilde{\calO} \rbr{\frac{d}{\epsilon \Delta^2}}$ 
&  $\tilde{\calO} \rbr{\frac{d^2}{\epsilon \Delta^2}}$ 
&  $\tilde{\calO} \rbr{\frac{d^2}{\epsilon \Delta^2}}$ \\ \hline
\caja{c}{c}{Streaming shift-and-invert\\(assuming close init. for $\lambda$)}
 & $\calO \rbr{\frac{d}{\epsilon \Delta^2}}$ 
&  $\calO \rbr{\frac{d^2}{\epsilon \Delta^2}}$ 
&  $\calO(d)$ \\
\hline
\end{tabular}
\end{table}\section{Conclusion}
\label{sec:conclusion}

In this paper, we have studied the sample complexity of population CCA for several
classes of input distributions, and proposed sample-efficient algorithms
for learning the first pair of canonical directions. While the original
problem is nonconvex, we exploit its structure as an eigenvalue problem, and analyze the statistical performance of the shift-and-invert power iterations. 

Based on the deflation/peeling scheme~\citep{Allen-ZhuLi16a, ZhuLi17a} for 
eigenvalue problems, our results shall be extended to extracting the top-k
canonical direction pairs. 
Our algorithms also apply to related eigenvalue problems in machine
learning, such as partial least squares~\citep{Chen_17a} and linear discriminant
analysis~\citep{BachJordan05a}, which 
are special versions of CCA with the population covariances being identity 
(\ie, $\Exx=\Eyy=\I$) and $\y$ being one-hot representations for class 
labels respectively. 
It is an interesting question if our general approach can be adapted to study the statistical performance of the kernel extension of CCA~\citep{Fukumiz_07a}. 
\section*{Acknowledgement}
Research partially supported by NSF BIGDATA award 1546462.

\appendix
\section{Auxiliary Lemmas}

\begin{lemma}\label{lem:population-obj-bound}
  The population canonical correlation is bounded by $1$, \ie,
  \begin{align*}
    \rho_1 = \sigma_1 \left( \T \right) \le 1.
  \end{align*}
\end{lemma}
\begin{proof}
  By the Cauchy-Schwarz inequality of random variables, we have 
  \begin{align*}
    \rho_1 = \bbE[ ({\uu^*}^\top \x) ({\vv^*}^\top \y) ]
    \le \sqrt{\bbE[ ({\uu^*}^\top \x)^2]} \cdot \sqrt{ \bbE[ ({\vv^*}^\top \y)^2] } 
    = \sqrt{ {\uu^*}^\top \Exx \uu } \cdot \sqrt{ {\vv^*}^\top \Eyy \vv }
    = 1 .
  \end{align*}
\end{proof}

\begin{lemma}[Distance between normalized vectors] \label{lem:normalized-difference}
  For two nonzero vectors $\aa, \b \in \bbR^{d}$, we have 
  \begin{align*}
    \norm{ \frac{\aa}{\norm{\aa}} - \frac{\b}{\norm{\b}}} \le 2 \frac{\norm{\aa-\b}}{\norm{\aa}}.
  \end{align*}
\end{lemma}
\begin{proof} By direct calculation, we have 
  \begin{align*}
    \norm{ \frac{\aa}{\norm{\aa}} - \frac{\b}{\norm{\b}}} & \le 
    \norm{ \frac{\aa}{\norm{\aa}} - \frac{\b}{\norm{\aa}}} +
    \norm{ \frac{\b}{\norm{\aa}} - \frac{\b}{\norm{\b}}} \\
    & =  \frac{ \norm{\aa - \b}}{\norm{\aa}} +
    \norm{\b} \cdot  \frac{ \abs{ \norm{\aa} - \norm{\b}}}{\norm{\aa} \norm{\b}}\\
    & \le \frac{ \norm{\aa - \b}}{\norm{\aa}} + \frac{ \norm{\aa - \b}}{\norm{\aa}} \\
    & = 2 \frac{\norm{\aa-\b}}{\norm{\aa}}
  \end{align*}
  where we have used the triangle inequality in the two inequalities.
\end{proof}

\begin{lemma}[Conversion from joint alignment to separate alignment] \label{lem:alignment-transfer}
Let $\eta\in \left(0,\frac{1}{4}\right)$. Consider the four nonzero vectors $\aa,\x \in \bbR^{d_x}$ and $\b,\y \in \bbR^{d_y}$ such that $\norm{\aa}=\norm{\b}=1$. If 
\begin{align} \label{e:joint-alignment}
\frac{1}{\sqrt{2}} \cdot \frac{\aa^\top \x + \b^\top \y}{\sqrt{\norm{\x}^2 + \norm{\y}^2}} \ge 1 - \eta,
\end{align}
we also have
  \begin{align*}
    \frac{1}{2} \left( \abs{ \frac{\aa^\top \x}{\norm{\x}} } + \abs{ \frac{\b^\top \y}{\norm{\y}} } \right) \ge 1 - 4 \eta.
  \end{align*}
\end{lemma}
\begin{proof} By the Cauchy-Schwarz inequality, we have
  \begin{align*}
\frac{\aa^\top \x + \b^\top \y}{\sqrt{\norm{\x}^2 + \norm{\y}^2}}
& = \frac{\aa^\top \x}{\norm{\x}} \cdot \frac{\norm{\x}}{\sqrt{\norm{\x}^2 + \norm{\y}^2}} + \frac{\b^\top \y}{\norm{\y}} \cdot \frac{\norm{\y}}{\sqrt{\norm{\x}^2 + \norm{\y}^2}} \\
& \le \sqrt{ \left( \frac{\aa^\top \x}{\norm{\x}} \right)^2 + 
\left( \frac{\b^\top \y}{\norm{\y}} \right)^2 }.
  \end{align*}
Thus according to~\eqref{e:joint-alignment}, we obtain
\begin{align*}
\left( \frac{\aa^\top \x}{\norm{\x}} \right)^2 + \left( \frac{\b^\top \y}{\norm{\y}} \right)^2
\ge 2 (1-\eta)^2 \ge 2 - 4 \eta.
\end{align*}
Since $\left( \frac{\b^\top \y}{\norm{\y}} \right)^2 \le 1$, this implies
\begin{align*}
\abs{\frac{\aa^\top \x}{\norm{\x}}} \ge \sqrt{1 - 4 \eta} \ge 1 - 4 \eta
\end{align*}
where the last step is due to the fact that $\sqrt{x}\ge x$ for $x \in (0,1)$. 
Similarly we have $\abs{\frac{\b^\top \y}{\norm{\y}}} \ge 1 - 4 \eta$. Then the theorem follows.
\end{proof}

\begin{lemma}\emph{\textbf{(Moment inequalities of sub-Gaussian and regular polynomial-tail random vectors)}}
\label{lem:sub-gaussian-moments}
Let $\z \in \bbR^d$ be isotropic and sub-Gaussian or regular polynomial-tail (see their definitions in Lemma~\ref{lem:probability}). Then for some constant $C^\prime > 0$, we have
\begin{align*}
\bbE \norm{\z}^2 \le d, \qquad \bbE \norm{\z}^4 \le C^\prime d^2, \qquad \bbE \abs{\q^\top \z}^4 \le C^\prime
\end{align*}
where $\q$ is any unit vector.
\end{lemma}
\begin{proof}
\textbf{Sub-Gaussian case\ }
The first bound is by $\bbE \norm{\z}^2=\bbE \trace{\z \z^\top}=\trace{I}=d$. To prove the second one, note that according to Theorem 2.1 in~\citet{Hsu_12b}, we have
\begin{align*}
\bbP \left( \norm{\z}^2 > C_1 (d+t) \right) < e^{-t}
\end{align*}
for all $t>0$. Therefore
\begin{align*}
\bbE \norm{\z}^4 &= \int_0^{\infty} \bbP \left( \norm{\z}^4 > s \right) ds \\
&= \int_0^{C_1^2 d^2} \bbP \left(\norm{\z}^4 > s \right) ds + \int_{C_1^2 d^2}^{\infty} \bbP \left(\norm{\z}^4 > s \right) ds \\
&\leq C_1^2 d^2 + \int_{C_1^2 d^2}^{\infty}\exp\left(-\left(\frac{\sqrt{s}}{C_1}-d\right)\right)ds \\
&\leq C^\prime d^2.
\end{align*}
Lastly,
\begin{align*}
\bbE \abs{\q^\top \z}^4 &= \int_0^{\infty}\bbP \left( \abs{\q^\top \z}^4 > s \right) ds \\
&\leq \int_0^{\infty} e^{-C \sqrt{s}}ds \\
&\leq C^\prime.
\end{align*}

\textbf{Regular polynomial-tail case\ } The first bound is still by $\bbE \norm{\z}^2=\bbE \trace{\z \z^\top} = \trace{I}=d$. 
When $r>1$, we have
\begin{align*}
\bbE \norm{\z}^4 &= \int_0^{\infty} \bbP \left( \norm{\z}^4 > s \right) ds \\
&\leq \int_0^{C^2 d^2} \bbP \left( \norm{\z}^4 > s \right) ds + \int_{C^2d^2}^{\infty} \bbP \left( \norm{\z}^4 > s \right) ds \\
&\leq C^2 d^2 + \int_{C^2 d^2}^{\infty} C s^{-\frac{1+r}{2}} ds \\
&\leq C^\prime d^2.
\end{align*}
To prove the last bound, take $\V =\q \q^\top$ in the definition of regular polynomial-tail random vectors, and then
\begin{align*}
\bbP\left( \abs{\q^\top \z}^2 > t \right) \leq C t^{-1-r},
\end{align*} 
for any $t>C$. We have
\begin{align*}
\bbE \abs{\q^\top \z}^4 & = \int_0^{\infty} \bbP \left( \abs{\q^\top \z}^4 > s \right) ds \\
&\leq \int_0^{C^2} \bbP \left(\abs{\q^\top \z}^4 > s \right) ds + \int_{C^2}^{\infty} \bbP \left( \abs{\q^\top \z}^4 > s \right) ds \\
&\leq C^2 + \int_{C^2}^{\infty} C s^{-\frac{1+r}{2}} ds \\
&\leq C^\prime.
\end{align*}
\end{proof}



\section{Proofs for Section~\ref{sec:intro}}

\subsection{Proof of Lemma~\ref{lem:transform-alignment-correlation}}
\begin{proof}
  Using the fact that $\frac{ {\uu}^\top \Exx \uu^*}{\norm{\Exx^{\frac{1}{2}} {\uu}}}$ and 
  $\frac{ {\vv}^\top \Eyy \vv^*}{\norm{\Eyy^{\frac{1}{2}} {\vv}}}$ are at most 1, the condition on alignment implies
  \begin{align*}
    \frac{ {\uu}^\top \Exx \uu^*}{\norm{\Exx^{\frac{1}{2}} {\uu}}} = \aa_1^\top \frac{\Exx^{\frac{1}{2}} \uu}{\norm{\Exx^{\frac{1}{2}} \uu}} \ge 1 - \frac{\eta}{4},\qquad 
    \frac{ {\vv}^\top \Eyy \vv^*}{\norm{\Eyy^{\frac{1}{2}} {\vv}}} = \b_1^\top \frac{\Eyy^{\frac{1}{2}} \vv}{\norm{\Eyy^{\frac{1}{2}} \vv}} \ge 1 - \frac{\eta}{4}.
  \end{align*}
  Since $\{\aa_i\}_{i=1}^r$ and $\{\b_i\}_{i=1}^r$ are orthonormal, we have
  \begin{align*}
    &\sum_{i=2}^r \left( \aa_i^\top \frac{\Exx^{\frac{1}{2}} \uu}{\norm{\Exx^{\frac{1}{2}} \uu}} \right)^2 \le 1 - \left(1-\frac{\eta}{4} \right)^2 \le \frac{\eta}{2}, \\
    & \sum_{i=2}^r \left( \b_i^\top \frac{\Eyy^{\frac{1}{2}} \vv}{\norm{\Eyy^{\frac{1}{2}} \vv}} \right)^2 \le 1 - \left(1-\frac{\eta}{4} \right)^2 \le \frac{\eta}{2}.
  \end{align*}
  Observe that
  \begin{align*}
    & \frac{\uu^\top \Exy \vv}{ \sqrt{\uu^\top \Exx \uu} \sqrt{\vv^\top \Eyy \vv} }
    = \frac{(\Exx^{\frac{1}{2}} \uu)^\top \T (\Eyy^{\frac{1}{2}} \vv)}{ \norm{\Exx^{\frac{1}{2}}\uu} \norm{\Eyy^{\frac{1}{2}} \vv}}
    = \sum_{i=1}^d \rho_i \left( \aa_i^\top \frac{\Exx^{\frac{1}{2}} \uu}{\norm{\Exx^{\frac{1}{2}} \uu}} \right) \left( \b_i^\top \frac{\Eyy^{\frac{1}{2}} \vv}{\norm{\Eyy^{\frac{1}{2}} \vv}} \right) \\
    \ge\ & \rho_1 \left( \aa_i^\top \frac{\Exx^{\frac{1}{2}} \uu}{\norm{\Exx^{\frac{1}{2}} \uu}} \right) \left( \b_1^\top \frac{\Eyy^{\frac{1}{2}} \vv}{\norm{\Eyy^{\frac{1}{2}} \vv}} \right) - \rho_2 \sqrt{ \sum_{i=2}^r \left(\aa_i^\top \frac{\Exx^{\frac{1}{2}} \uu}{\norm{\Exx^{\frac{1}{2}} \uu}} \right)^2 }
    \sqrt{\sum_{i=2}^r \left(\b_i^\top \frac{\Eyy^{\frac{1}{2}} \vv}{\norm{\Eyy^{\frac{1}{2}} \vv}}\right)^2} \\
    \ge\ & \rho_1 \left(1- \frac{\eta}{4}\right)^2 - \rho_1 \cdot \frac{\eta}{2} \ge \rho_1 \left( 1 - \eta \right)
  \end{align*}
  where we have used the Cauchy-Schwarz inequality in the first inequality.
\end{proof}

\section{Proofs for Section~\ref{sec:erm}}

\subsection{Proof of Lemma~\ref{lem:probability}}

\begin{proof} \textbf{Sub-Gaussian/regular polynomial-tail cases\ } Consider the random variable $\z$ defined in~\eqref{e:distributions-z}, and draw i.i.d. samples $\z_1,\dots,\z_n$ of $\z$. 
It is known that when the sample size $n$ is large enough (as specified in the lemma), we have
\begin{align*}
\norm{\frac{1}{N} \sum_{i=1}^N \z_i \z_i^\top - \I } \le \frac{\nu}{2}
\end{align*}
with high probability for the sub-Gaussian class~\citep{Vershy12a} and for the regular polynomial-tail class~\citep{SrivasVershy13a}, given $N > C^\prime \frac{d}{\nu^2}$ and $N \ge C^\prime \frac{d}{\nu^{2 (1+r^{-1})}}$ respectively.

We then turn to bounding the error in each covariance matrix. We note that the covariance of $\f:=\left[ \begin{array}{c} \Exx^{-\frac{1}{2}} \x \\  \Eyy^{-\frac{1}{2}} \y \end{array} \right]$ is $
\bXi = \left[ \begin{array}{cc} \I & \T \\ \T^\top & \I \end{array} \right]
$ 
with $\norm{\bXi} = 1 + \rho_1 \le 2$ (since the eigenvalues of
$\bXi$ are of the form $1 \pm \sigma_i (\T)$). On the other hand, define
\begin{align*}
\bPhi:=\left[\begin{array}{cc}\Exx & \\ & \Eyy\end{array}\right]^{-\frac{1}{2}}
   \left[\begin{array}{cc}\Exx &  \Exy \\ \Exy^\top & \Eyy\end{array}\right]^{\frac{1}{2}}
\qquad\text{satisfying}\quad
\bPhi \bPhi^\top = \bXi
\end{align*}
and we have $\f = \bPhi \z$ by the definition of $\z$. Furthermore, $\f_i = \bPhi \z_i$, $i=1,\dots,N$ are i.i.d. samples of $\f$. Therefore, it holds that
\begin{gather*}
\norm{\left[
\begin{array}{cc}
\frac{1}{N} \sum_{i=1}^N \Exx^{-\frac{1}{2}} \x_i \x_i^\top \Exx^{-\frac{1}{2}} - \I & 
\frac{1}{N} \sum_{i=1}^N \Exx^{-\frac{1}{2}} \x_i \y_i^\top \Eyy^{-\frac{1}{2}} - \T \\
\frac{1}{N} \sum_{i=1}^N \Eyy^{-\frac{1}{2}} \y_i \x_i^\top \Exx^{-\frac{1}{2}} - \T^\top &
\frac{1}{N} \sum_{i=1}^N \Eyy^{-\frac{1}{2}} \y_i \y_i^\top \Eyy^{-\frac{1}{2}} - \I 
\end{array}
\right]}
= \norm{ \frac{1}{N} \sum_{i=1}^N \f_i \f_i^\top - \bXi }  \\
= \norm{ \bPhi \left(\frac{1}{N} \sum_{i=1}^N \z_i \z_i^\top - \I\right) \bPhi^\top } 
\le \norm{\bPhi \bPhi^\top} \cdot \norm{ \frac{1}{N} \sum_{i=1}^N \z_i \z_i^\top - \I  } 
= \norm{\bXi} \cdot \norm{ \frac{1}{N} \sum_{i=1}^N \z_i \z_i^\top - \I  } \le \nu.
\end{gather*}
Since the norm of each block is bounded by the norm of the entire matrix, we conclude that the error in estimating each covariance matrix is bounded by $\nu$, as required by Proposition~\ref{prop:probabilistic-property}.

\begin{remark} \label{rmk:data-norm-bound}
In view of Lemma~\ref{lem:sub-gaussian-moments} and the proof technique here, for the sub-Gaussian/regular polynomial-tail cases, the bound of $\norm{\z}^2$ leads to a bound for $\norm{\x}^2$ and $\norm{\y}^2$: 
\begin{align*}
\bbE ( \norm{\x}^2 + \norm{\y}^2) \le \norm{\Exx} \cdot \bbE \norm{\Exx^{-\frac{1}{2}} \x}^2 + \norm{\Eyy} \cdot \bbE \norm{\Eyy^{-\frac{1}{2}} \y}^2 
\le \bbE \norm{\f}^2 
\le 2 \bbE \norm{\z}^2 
\le C d
\end{align*}
for some constant $C>0$, where we have used Assumption~\ref{assump:bounded-covariances} in the second inequality. 
And similarly, we have
\begin{align*}
\bbE ( \norm{\x}^2 + \norm{\y}^2)^2 \le \bbE \norm{\f}^4
\le 4 \bbE \norm{\z}^4
\le C^\prime d^2 
\end{align*}
for some constant $C^\prime > 0$.
\end{remark} 

\textbf{Bounded case\ } 
Consider the joint covariance matrix 
\begin{align*}
  \left[\begin{array}{cc}\Exx & \Exy \\ \Exy^\top & \Eyy\end{array}\right] \in \bbR^{d \times d}
\end{align*}
which has eigenvalue bounded by $2$ due to the assumption that $\norm{\x}^2 + \norm{\y}^2 \le 2$. 
Applying~\citet[Corollary~5.52]{Vershy12a}, we obtain that 
\begin{align} \label{e:probability-joint-covariance-bernstein}
\norm{ \left[\begin{array}{cc} \Sxx & \Sxy \\ \Sxy^\top & \Syy\end{array}\right] - \left[\begin{array}{cc} \Exx & \Exy \\ \Exy^\top & \Eyy\end{array}\right] } \le \nu^\prime
\end{align}
with probability at least $1 - d^{-t^2}$ when $N \ge C (t/\nu^\prime)^2 \log d$ for some constant $C>0$. Setting the failure probability $\delta=d^{-t^2}$ gives $t^2 = \frac{\log \frac{1}{\delta}} { \log d }$, and thus we require $N \ge C \frac{1}{ {\nu^\prime}^2} \log \frac{1}{\delta}$ for $1 - \delta$ success probability.


Due to the block structure of the joint covariance matrix,~\eqref{e:probability-joint-covariance-bernstein} implies 
\begin{align*}
\norm{\Sxy - \Exy} \le \nu^\prime, \qquad  \norm{\Sxx - \Exx} \le \nu^\prime , \qquad 
\norm{\Syy - \Eyy} \le \nu^\prime
\end{align*}
hold simultaneously. 

Now, to satisfy the first inequality of~\eqref{e:property-probability}, observe that 
\begin{align*}
\norm{\Exx^{-\frac{1}{2}} \Sxx \Exx^{-\frac{1}{2}} - \I } 
& = \norm{\Exx^{-\frac{1}{2}} (\Sxx - \Exx) \Exx^{-\frac{1}{2}} } \\
& \le \norm{\Exx^{-\frac{1}{2}}} \cdot \norm{\Sxx - \Exx} \cdot \norm{\Exx^{-\frac{1}{2}} } \\
& \le \norm{\Sxx - \Exx} / \gamma
\end{align*}
where we have used the assumption that $\sigma_{\min} (\Exx) \ge \gamma$ in the last inequality. 
Therefore, we obtain $\norm{\Exx^{-\frac{1}{2}} \Sxx \Exx^{-\frac{1}{2}} - \I } \le \nu$ by setting $\nu^\prime = \gamma \nu$ in~\eqref{e:probability-joint-covariance-bernstein}, and this yields the $N_0 (\nu)$ chosen in the lemma. 
The other two inequalities of~\eqref{e:property-probability} can be obtained analogously. 
\end{proof}

\subsection{Proof of Lemma~\ref{lem:perturbation-matrix-square-root}}

\begin{proof}
This result can be derived from the main result of \citet{Mathias97a} with a bit
of detective work, which is needed to understand the higher order error
term. 
As in~\citet{Mathias97a}, assume without loss of generality that 
$\HH=\diag{\lambda_1,\dots,\lambda_d}$ is diagonal.
In the proof of Theorem~2,~\citet{Mathias97a} applied the
Daleckii-Krein formula in his equation (4) with the $\calO$-notation, which can be rephrased
as (see also~\citet{Carlsson_18e}[Theorem~2.1]): for $\zeta$ in a
neighborhood of $0$, it holds
\begin{align*}
\varepsilon :=
\norm{ (\HH+\zeta \bTheta)^{\frac{1}{2}} - \HH^{\frac{1}{2}} - \zeta \sbr{\frac{\lambda_i^{\frac{1}{2}}
  \lambda_j^{\frac{1}{2}}}{\lambda_i^{\frac{1}{2}} +
  \lambda_j^{\frac{1}{2}}}}_{i,j=1}^d \circ (\HH^{-\frac{1}{2}} \bTheta
  \HH^{-\frac{1}{2}})}  = \calO (\zeta^2)
\end{align*}
where $\circ$ denotes elementwise (Hadamard) multiplication. 

To locate the neighborhood of $\zeta$ for which the
above is true, we apply the matrix mean value
theorem~\citep{Gekel81a} with the second order derivative of matrix square
root~\citep[Theorem~6.6.30 and page 549]{HornJohnson91a} to obtain:
\begin{align}
\varepsilon \le & \frac{\zeta^2}{2} \max_{\tau \in [0, \zeta]} \bigg\| \U(\tau) \bigg( \sum_{k=1}^d 
\sbr{\frac{1}{\sqrt{\lambda_i(\tau)}+\sqrt{\lambda_j(\tau)}} \cdot
  \frac{1}{\sqrt{\lambda_i(\tau)}+\sqrt{\lambda_k(\tau)}} \cdot
  \frac{1}{\sqrt{\lambda_j(\tau)}+\sqrt{\lambda_k(\tau)}}}_{i,j=1}^d
 \nonumber \\ \label{e:matrix-sqrt-hessian}
&\qquad \circ \sbr{\c_k(\tau) \c_k^\top(\tau)} \bigg) \U(\tau)^\top \bigg\|
\end{align}
where $\HH + \tau \bTheta = \U(\tau) \cdot \diag{\lambda_1(\tau),\dots,\lambda_d (\tau)} \cdot
\U(\tau)^\top$ is the eigenvalue decomposition of the perturbation of $\HH$, and $\c_k(\tau)$ is the
$k$-th column of $\X(\tau) = \U(\tau)^\top \bTheta \U(\tau)$.

Define $\Z(\tau) =
\sbr{\frac{1}{\sqrt{\lambda_i(\tau)}+\sqrt{\lambda_j(\tau)}}}_{i,j=1}^d$. The summation
enclosed in $()$ of the right hand side of~\eqref{e:matrix-sqrt-hessian} can be
written as $\Z(\tau) \circ (\Z(\tau) \circ \X(\tau))^2$. Thus continuing
from~\eqref{e:matrix-sqrt-hessian} yields
\begin{align*}
\varepsilon \le \frac{\zeta^2}{2} \cdot \max_{\tau  \in [0, \zeta]} \norm{\Z(\tau) \circ (\Z(\tau) \circ \X(\tau))^2}.
\end{align*}
On the one hand, by the assumption that $\norm{\HH} \le \sigma_{\max}$, we have
\begin{align} \label{e:matrix-sqrt-perturb-to-H}
\norm{\X(\tau)} = \norm{\bTheta} = \norm{\HH^{\frac{1}{2}}
  (\HH^{-\frac{1}{2}} \bTheta \HH^{-\frac{1}{2}}) \HH^{\frac{1}{2}}} \le
  \norm{\HH} \cdot \norm{\HH^{-\frac{1}{2}} \bTheta \HH^{-\frac{1}{2}}} \le \sigma_{\max}. 
\end{align}
On the other hand, the matrix $\Z(\tau)$ is positive semidefinite (see
\citealp[Problem 9, page~348]{HornJohnson91a}). Using the fact that $\norm{\A \circ \B} \le (\max_{i} \A_{ii}) \cdot
\norm{\B}$ for positive semidefinite $\A$ and Hermitian $\B$~\citep[Theorem~5.5.18]{HornJohnson91a}, we conclude 
\begin{align*}
\varepsilon \le \frac{\zeta^2 \sigma_{\max}^2}{2} \sbr{\max_{\tau  \in [0, \zeta]} \max_i \frac{1}{2 \sqrt{\lambda_i (\tau)}}}^3.
\end{align*}
Let $\zeta \le \frac{3}{4} \sigma_{\max}^{-1} \sigma_{\min}$. Then in view
of~\eqref{e:matrix-sqrt-perturb-to-H} and the
Weyl's inequality, $\lambda_i (\tau) \ge \frac{\sigma_{\min}}{4}$ for all
$\tau\in[0,\zeta]$, and we have $\varepsilon \le \frac{1}{2} \zeta^2
\sigma_{\max}^2 \sigma_{\min}^{-\frac{3}{2}}$. 

To sum up, we have shown so far the following first order approximation:
for certain error matrix $\calE \in \bbR^{d\times d}$, it holds
\begin{align*}
(\HH+\zeta \bTheta)^{\frac{1}{2}} = \HH^{\frac{1}{2}} + \zeta \sbr{\frac{\lambda_i^{\frac{1}{2}}
  \lambda_j^{\frac{1}{2}}}{\lambda_i^{\frac{1}{2}} +
  \lambda_j^{\frac{1}{2}}}}_{i,j=1}^d \circ (\HH^{-\frac{1}{2}} \bTheta
  \HH^{-\frac{1}{2}}) + \calE \qquad\text{where} \quad \norm{\calE} \le
  \frac{1}{2} \zeta^2 \sigma_{\max}^2 \sigma_{\min}^{-\frac{3}{2}}.
\end{align*}
Consequently, we have 
\begin{align*}
(\HH+\zeta \bTheta)^{\frac{1}{2}} \HH^{-\frac{1}{2}}  - \I  
= \zeta \sbr{\frac{\lambda_i^{\frac{1}{2}}}{\lambda_i^{\frac{1}{2}} +
  \lambda_j^{\frac{1}{2}}}}_{i,j=1}^d  \circ (\HH^{-\frac{1}{2}} \bTheta
  \HH^{-\frac{1}{2}}) + \calE \HH^{-\frac{1}{2}}.
\end{align*}
\citet{Mathias97a} showed that the norm of the first term on the right
hand size is of the order $\calO(\log d \cdot \zeta)$. Combining this with the fact that 
$\norm{\calE \HH^{-\frac{1}{2}}} \le \frac{1}{2} \zeta^2 \sigma_{\max}^2
\sigma_{\min}^{-2}$, we conclude that $\norm{\calE \HH^{-\frac{1}{2}}} =\calO
(\zeta)$ for $\zeta=\calO(\sigma_{\max}^{-2} \sigma_{\min}^2)$ and the lemma follows.
\end{proof}

\subsection{Proof of Lemma~\ref{lem:approx-error-erm}}
\begin{proof}
In view of the Weyl's inequality, we have
  \begin{align} \label{e:weyl}
    \abs{\widehat{\rho}_1 - \rho_1} \le \norm{ \widehat{\T} - \T } = \norm{ \Sxx^{-\frac{1}{2}} \Sxy \Syy^{-\frac{1}{2}} - \Exx^{-\frac{1}{2}} \Exy \Eyy^{-\frac{1}{2}} }.
  \end{align}
  For the right hand side of~\eqref{e:weyl}, we have the following decomposition
  \begin{align}
    & \Sxx^{-\frac{1}{2}} \Sxy \Syy^{-\frac{1}{2}} - \Exx^{-\frac{1}{2}} \Exy \Eyy^{-\frac{1}{2}} \nonumber \\ \label{e:sample-three-terms}
    = & \left( \Sxx^{-\frac{1}{2}} - \Exx^{-\frac{1}{2}} \right) \Sxy \Syy^{-\frac{1}{2}} + 
    \Exx^{-\frac{1}{2}} \left( \Sxy - \Exy \right) \Syy^{-\frac{1}{2}} + 
    \Exx^{-\frac{1}{2}}  \Exy \left( \Syy^{-\frac{1}{2}} - \Eyy^{-\frac{1}{2}} \right).
  \end{align} 
  By the equality
  \begin{align*} 
    \A^{-\frac{1}{2}} - \B^{-\frac{1}{2}} = \B^{-\frac{1}{2}} \left( \B^{\frac{1}{2}} - \A^{\frac{1}{2}} \right) \A^{-\frac{1}{2}},
  \end{align*}
  the first term of the RHS of~\eqref{e:sample-three-terms} becomes
  \begin{align*}
    \left( \Sxx^{-\frac{1}{2}} - \Exx^{-\frac{1}{2}} \right) \Sxy \Syy^{-\frac{1}{2}}
    = \Exx^{-\frac{1}{2}} \left( \Exx^{\frac{1}{2}} - \Sxx^{\frac{1}{2}} \right)
    \Sxx^{-\frac{1}{2}} \Sxy \Syy^{-\frac{1}{2}}.
  \end{align*} 
When $\norm{\Exx^{-\frac{1}{2}} \Sxx \Exx^{-\frac{1}{2}} - \I } \le \nu$,
according to Lemma~\ref{lem:perturbation-matrix-square-root}, we have (by
making the identification that $\HH = \Exx$,
$\zeta=\norm{\Exx^{-\frac{1}{2}} (\Sxx - \Exx) \Exx^{-\frac{1}{2}}} $,
and $\bTheta = (\Sxx - \Exx) /\norm{\Exx^{-\frac{1}{2}} (\Sxx - \Exx) \Exx^{-\frac{1}{2}}} $)
\begin{align*}
\norm{\Exx^{-\frac{1}{2}} \left( \Exx^{\frac{1}{2}} - \Sxx^{\frac{1}{2}} \right)} \le C_d \cdot \nu
 \end{align*}
 for $\nu \le \frac{3}{4} \gamma^2$. 
Combining with the fact that $\norm{\Sxx^{-\frac{1}{2}} \Sxy \Syy^{-\frac{1}{2}}} \le 1$, we have
  \begin{align*}
    \norm{ \left( \Sxx^{-\frac{1}{2}} - \Exx^{-\frac{1}{2}} \right) \Sxy \Syy^{-\frac{1}{2}} }
    \le C_d \cdot \nu. 
  \end{align*} 
  A similar bound can be obtained for the third term
  of~\eqref{e:sample-three-terms}. Observe that when
  $\norm{\Eyy^{-\frac{1}{2}} \Syy \Eyy^{-\frac{1}{2}} - \I } \le \nu < 1$,
  we have $\norm{\Syy - \Eyy} \le \nu$ and all eigenvalues of $\Syy$ lie
  in $[\gamma - \nu, 1 + \nu]$. Additionally, $\Eyy^{-\frac{1}{2}} \Syy
  \Eyy^{-\frac{1}{2}}$ is invertible, and all eigenvalues of $\Syy^{-\frac{1}{2}}
  \Eyy \Syy^{-\frac{1}{2}}$ lie in $[\frac{1}{1+\nu}, \frac{1}{1-\nu}]$, impling that $\norm{\Syy^{-\frac{1}{2}} \Eyy
    \Syy^{-\frac{1}{2}} - \I} \le \frac{\nu}{1-\nu}$. According to
  Lemma~\ref{lem:perturbation-matrix-square-root}, we have (by making the
  identification that $\HH = \Syy$, $\zeta=\norm{\Syy^{-\frac{1}{2}} (\Syy - \Eyy)  \Syy^{-\frac{1}{2}}}$, and
  $\bTheta = (\Syy - \Eyy) / \norm{\Syy^{-\frac{1}{2}} (\Syy - \Eyy)  \Syy^{-\frac{1}{2}}} $)  
\begin{align} \label{e:error-decomposition-matrix-square-root}
  \norm{\left(\Eyy^{\frac{1}{2}} - \Syy^{\frac{1}{2}} \right) \Syy^{-\frac{1}{2}} } \le \frac{C_d \cdot \nu}{1-\nu}
\end{align}
for $\nu \le \frac{3}{4} \left(\frac{\gamma - \nu}{1 + \nu}\right)^2$,
which is satisfied for $\nu \le \frac{1}{4} \gamma^2$. 
Therefore, we can bound the third term of~\eqref{e:sample-three-terms} as
  \begin{align*}
    \norm{ \Exx^{-\frac{1}{2}}  \Exy \left( \Syy^{-\frac{1}{2}} - \Eyy^{-\frac{1}{2}} \right) } & = \norm{ \Exx^{-\frac{1}{2}}  \Exy \Eyy^{-\frac{1}{2}} \left(\Eyy^{\frac{1}{2}} - \Syy^{\frac{1}{2}} \right) \Syy^{-\frac{1}{2}} } \\
    & \le \norm{ \Exx^{-\frac{1}{2}}  \Exy \Eyy^{-\frac{1}{2}} } \cdot \norm{ \left(\Eyy^{\frac{1}{2}} - \Syy^{\frac{1}{2}} \right) \Syy^{-\frac{1}{2}} } \\
    & \le \frac{C_d \cdot \nu}{1-\nu} \le 2 C_d \cdot \nu
  \end{align*}
where we have used the fact that $\norm{ \Exx^{-\frac{1}{2}}  \Exy \Eyy^{-\frac{1}{2}}}\le 1$ by Lemma~\ref{lem:population-obj-bound}.

  For the second term of~\eqref{e:sample-three-terms}, we have by assumption that 
  \begin{align*}
    \norm{\Exx^{-\frac{1}{2}} \left( \Sxy - \Exy \right) \Syy^{-\frac{1}{2}}}
    \le \nu.
  \end{align*}
  Applying the triangle inequality, we obtain from~\eqref{e:sample-three-terms} that  
  \begin{align} \label{e:sample-three-terms-2}
    \norm{ \Sxx^{-\frac{1}{2}} \Sxy \Syy^{-\frac{1}{2}} - \Exx^{-\frac{1}{2}} \Exy \Eyy^{-\frac{1}{2}} }
    \le 
    4 C_d \cdot \nu.
  \end{align}
To sum up, it suffices to set $\nu = \frac{\epsilon^\prime}{4 C_d}$ to ensure $\norm{\widehat{\T} - \T} \le \epsilon^\prime$, and this yields the desired sample complexity.
\end{proof}

\subsection{Proof of Theorem~\ref{thm:approx-direction-erm}}

\begin{proof}
  Apply Lemma~\ref{lem:approx-error-erm} with $\epsilon^\prime =
  \frac{\sqrt{\epsilon} \Delta}{4}$ where $\epsilon \in (0,1)$ is the
  desired accuracy in Theorem~\ref{thm:approx-direction-erm}. 
Since $\norm{\T - \hat{\T}} \le \epsilon^\prime < \frac{\Delta}{4}$, the
eigenvalues of $\hat{\T}$ are within $\frac{\Delta}{4}$ of those of $\T$
due to Weyl's inequality, so there exists a positive singular gap of
$\frac{\Delta}{2}$ for the empirical estimate $\hat{\T}$, whose first pair
of singular vectors is unique. 
In view of the off-diagonal structure of $\widehat{\C}$, we observe that
$\norm{\C - \hat{\C}} = \norm{\T - \hat{\T}} \le \epsilon^\prime$ and that the top eigenvector
of $\hat{\C}$ is unique. 
Then with the number of samples given in the
theorem ensuring the $\epsilon^\prime$ perturbation, according to the Davis-Kahan $\sin \theta$
theorem~\citep{DavisKahan70a}, the top eigenvectors of $\C$ and $\widehat{\C}$ are well aligned:
  \begin{align} \label{e:sin-erm-population}
    \sin^2 \theta \le \frac{\norm{\C - \widehat{\C}}^2}{\Delta^2} \le
    \frac{{\epsilon^\prime}^2}{\Delta^2} = \frac{\epsilon}{16} 
  \end{align}
  where $\theta$ is the angle between the top eigenvector of $\C$ and that of $\widehat{\C}$. 
  This is equivalent to 
\begin{align} \label{e:erm-fix-normalization}
\norm{ \Exx^{\frac{1}{2}} \uu^* - \Sxx^{\frac{1}{2}} \widehat{\uu} }^2 
+ \norm{ \Eyy^{\frac{1}{2}} \vv^* - \Syy^{\frac{1}{2}} \widehat{\vv} }^2 \le \frac{\epsilon}{8}
\end{align}
and so $\max \left( \norm{ \Exx^{\frac{1}{2}} \uu^* - \Sxx^{\frac{1}{2}} \widehat{\uu} }^2,\, \norm{ \Eyy^{\frac{1}{2}} \vv^* - \Syy^{\frac{1}{2}} \widehat{\vv} }^2 \right) \le \frac{\epsilon}{8}$.

  In the rest of the proof, we fix the issue of incorrect normalization of $(\widehat{\uu}, \widehat{\vv})$. 
Recall we have shown in the proof of Lemma~\ref{lem:approx-error-erm} that (see \eg,~\eqref{e:error-decomposition-matrix-square-root})
\begin{align*}
\norm{ \I - \Exx^{\frac{1}{2}} \Sxx^{-\frac{1}{2}} } \le \epsilon^\prime \le \frac{\sqrt{\epsilon}}{4}.
\end{align*}

Consequently, we have
\begin{align*}
\norm{ \Exx^{\frac{1}{2}} \uu^* - \Exx^{\frac{1}{2}} \widehat{\uu} }^2
& = \norm{ \Exx^{\frac{1}{2}} \uu^* - \left( \Exx^{\frac{1}{2}} \Sxx^{-\frac{1}{2}} \right)  \left(\Sxx^{\frac{1}{2}} \widehat{\uu} \right) }^2 \\
& \le \left( \norm{ \Exx^{\frac{1}{2}} \uu^* - \Sxx^{\frac{1}{2}} \widehat{\uu} }  + \norm{ \left( \I - \Exx^{\frac{1}{2}} \Sxx^{-\frac{1}{2}} \right)  \Sxx^{\frac{1}{2}} \widehat{\uu}  } \right)^2 \\
& \le  2  \norm{ \Exx^{\frac{1}{2}} \uu^* - \Sxx^{\frac{1}{2}} \widehat{\uu} }^2  + 2 \norm{ \I - \Exx^{\frac{1}{2}} \Sxx^{-\frac{1}{2}} }^2  \\
& \le \frac{\epsilon}{4}  + \frac{\epsilon}{8} \\
& \le \frac{\epsilon}{2}
\end{align*}
where we have used the facts that $(x+y)^2 \le 2 x^2 + 2 y^2$ and $\norm{\Sxx^{\frac{1}{2}} \widehat{\uu}} = 1$ in the second inequality. 

 According to Lemma~\ref{lem:normalized-difference}, we then have
  \begin{align*}
    \norm{ \frac{\Exx^{\frac{1}{2}} \uu^* }{\norm{\Exx^{\frac{1}{2}} \uu^* }} - \frac{\Exx^{\frac{1}{2}} \widehat{\uu}}{\norm{\Exx^{\frac{1}{2}} \widehat{\uu}}} }^2
    \le \frac{4 \norm{ \Exx^{\frac{1}{2}} \uu^* - \Exx^{\frac{1}{2}} \widehat{\uu}}^2}{\norm{\Exx^{\frac{1}{2}} \uu^* }^2} 
    = 4 \norm{ \Exx^{\frac{1}{2}} \uu^* - \Exx^{\frac{1}{2}} \widehat{\uu}}^2 
    \le 2 \epsilon
  \end{align*}
and thus the alignment between these two vectors is
  \begin{align*}
\frac{ \widehat{\uu}^\top \Exx \uu^*}{\norm{\Exx^{\frac{1}{2}} \widehat{\uu}}} = 
1 - \frac{1}{2} \norm{ \frac{\Exx^{\frac{1}{2}} \uu^* }{\norm{\Exx^{\frac{1}{2}} \uu^* }} - \frac{\Exx^{\frac{1}{2}} \widehat{\uu}}{\norm{\Exx^{\frac{1}{2}} \widehat{\uu}}} }^2 
\ge 1 - \epsilon .
  \end{align*}

A similar bound is obtained for $\widehat{\vv}$:
  \begin{align*}
\frac{ \widehat{\vv}^\top \Eyy \vv^*}{\norm{\Eyy^{\frac{1}{2}} \widehat{\vv}}} \ge 1 - \epsilon .
  \end{align*}
Averaging the above two inequalities yields the desired result.  
Requiring that  $\epsilon^\prime = \frac{\sqrt{\epsilon} \Delta}{4} \le
C_d \gamma^2$ as in Corollary~\ref{cor:sample-complexity-canoncorr} leads
to the extra condition that $\epsilon \le \frac{16 C_d^2 \gamma^4}{\Delta^2}$.
\end{proof}

\section{Proofs for Section~\ref{sec:stochastic-opt}}

\subsection{Proof of Lemma~\ref{lem:progress-of-matrix-vector-multiplication}}
\begin{proof}
  If we obtain an approximate solution $\w_{t+1}$ to~\eqref{e:lsq}, such that  $f_{t+1} (\w_{t+1}) - f_{t+1} (\w_{t+1}^*) = \epsilon_t (\w_t^\top \widehat{\B} \w_t) $, it holds that 
  \begin{align*}
    \epsilon_t \norm{\widehat{\B}^{\frac{1}{2}} \w_t}^2 & = \frac{1}{2} \left(\w_{t+1} - \w_{t+1}^* \right)^\top  \widehat{\A}_{\lambda} \left(\w_{t+1} - \w_{t+1}^* \right) \\
    & = \frac{1}{2} \left(\widehat{\B}^{\frac{1}{2}} \w_{t+1} - \widehat{\B}^{\frac{1}{2}} \w_{t+1}^* \right)^\top \widehat{\B}^{-\frac{1}{2}}  \widehat{\A}_{\lambda}  \widehat{\B}^{-\frac{1}{2}} \left(\widehat{\B}^{\frac{1}{2}} \w_{t+1} - \widehat{\B}^{\frac{1}{2}} \w_{t+1}^* \right) \\
    & = \frac{1}{2} \left( \rr_{t+1} - \rr_{t+1}^* \right)^\top \widehat{\M}_{\lambda}^{-1} \left( \rr_{t+1} - \rr_{t+1}^* \right) = \frac{1}{2} \normhatM{\rr_{t+1} - \rr_{t+1}^*}^2,
  \end{align*}
  or equivalently
  \begin{align*}
    \normhatM{\rr_{t+1} - \rr_{t+1}^*} = \sqrt{2 \epsilon_t } \cdot \norm{\rr_t}.
  \end{align*}
  Note that our choice of $\epsilon_t$ is also invariant to the length of $\rr_t$ (or whether normalization is performed).

  For the exact solution to the linear system, we have
  \begin{align*}
    \rr_{t+1}^* = \widehat{\M}_{\lambda} \rr_{t} = \norm{\rr_{t}} \sum_{i=1}^d \beta_i \xi_{ti} \p_i .
  \end{align*}
  As a result, we can bound the numerator and denominator of $G(\rr_{t+1})$ respectively:
  \begin{align*}
    \normhatM{\PP_{\perp} \frac{\rr_{t+1}}{\norm{\rr_{t+1}}} } & \le \frac{1}{\norm{\rr_{t+1}}}  \left( \normhatM{\PP_{\perp} \rr_{t+1}^* } + \normhatM{\PP_{\perp} \left( \rr_{t+1} - \rr_{t+1}^* \right) } \right) \\
    & \le \frac{1}{\norm{\rr_{t+1}}} \left( \normhatM{\PP_{\perp} \rr_{t+1}^* } + \normhatM{ \rr_{t+1} - \rr_{t+1}^* } \right) \\
    & = \frac{\norm{\rr_{t}}}{\norm{\rr_{t+1}}} \left( \sqrt{\sum_{i=2}^d \beta_i \xi_{ti}^2}  + \sqrt{2 \epsilon_t } \right),
  \end{align*}
  and 
  \begin{align*}
    \normhatM{\PP_{\parallel} \frac{\rr_{t+1}}{\norm{\rr_{t+1}}} } & \ge \frac{1}{\norm{\rr_{t+1}}} \left( \normhatM{\PP_{\parallel} \rr_{t+1}^* } - \normhatM{\PP_{\parallel} \left( \rr_{t+1} - \rr_{t+1}^* \right) } \right) \\
    & \ge \frac{1}{\norm{\rr_{t+1}}} \left(  \normhatM{\PP_{\parallel} \rr_{t+1}^* } - \normhatM{ \rr_{t+1} - \rr_{t+1}^* } \right) \\
    & = \frac{\norm{\rr_{t}}}{\norm{\rr_{t+1}}} \left( \sqrt{\beta_1 \xi_{t1}^2} - \sqrt{2 \epsilon_t} \right).
  \end{align*}

  Consequently, we have
  \begin{align*}
    G(\rr_{t+1}) & \le \frac{\sqrt{\sum_{i=2}^d \beta_i \xi_{ti}^2} + \sqrt{2 \epsilon_t }}{ \sqrt{\beta_1 \xi_{t1}^2} - \sqrt{2 \epsilon_t }}
    \le \frac{\beta_2 \sqrt{\sum_{i=2}^d \xi_{ti}^2 / \beta_i} + \sqrt{2 \epsilon_t} }{\beta_1 \sqrt{\xi_{t1}^2 / \beta_1} - \sqrt{2 \epsilon_t}} \\ 
    & = G(\rr_{t}) \cdot \frac{\beta_2  + \frac{\sqrt{2 \epsilon_t}}{\sqrt{\sum_{i=2}^d \xi_{ti}^2 / \beta_i}}}{\beta_1  - \frac{\sqrt{2 \epsilon_t}}{\sqrt{\xi_{t1}^2 / \beta_1}}}.
  \end{align*}

  As long as $\sqrt{2 \epsilon_t} \le \min \left( \sqrt{\sum_{i=2}^d \xi_{ti}^2 / \beta_i},\ \sqrt{\xi_{t1}^2 / \beta_1}  \right) \cdot \frac{\beta_1 - \beta_2}{4}$, \ie, 
  \begin{align*} 
    \epsilon_t \le  \min \left( {\sum_{i=2}^d \xi_{ti}^2 / \beta_i},\ {\xi_{t1}^2 / \beta_1} \right) \cdot  \frac{\left( \beta_1 - \beta_2 \right)^2}{32} ,
  \end{align*}
  we are guaranteed that
  \begin{align*}
    G(\rr_{t+1}) \le G(\rr_{t}) \cdot \frac{\beta_1 + 3 \beta_2}{3 \beta_1 + \beta_2}.
  \end{align*}
  Substituting in $\beta_i = \frac{1}{\lambda - \widehat{\rho}_i}$ with $\lambda - \widehat{\rho}_1 \le \widehat{\Delta}$, we obtain that 
  \begin{align*} 
    \frac{\beta_1 + 3 \beta_2}{3 \beta_1 + \beta_2} \le \frac{5}{7} < 1.
  \end{align*}
  This means that if $\eqref{e:epsilon-t}$ holds for each least squares problem, the sequence $\{G(\rr_{t})\}_{t=0,\dots}$ decreases (at least) at a constant geometric rate of $\frac{5}{7}$. Therefore, the number of inexact matrix-vector multiplications $T$ needed to achieve $\abs{\sin \theta_T} \le \eta$ is $\log_{\frac{7}{5}} \left( \frac{G(\rr_0)}{\eta} \right)$.
\end{proof}

\subsection{Bounding the initial error for each least squares}
\label{sec:erm-least-squares-initial-error}

We can minimize the initial suboptimality for the least squares problem $f_{t+1}$ for reducing the time complexity of its solver. It is natural to use an initialization of the form $\alpha \w_{t}$, a scaled version of the previous iterate, which gives the following  objective
\begin{align*}
  f_{t+1} (\alpha \w_{t}) = \frac{(\w_{t}^\top \widehat{\A}_{\lambda} \w_{t})}{2} \alpha^2 - (\w_{t}^\top \widehat{\B} \w_{t}) \alpha .
\end{align*}
This is a quadratic function of $\alpha$, and minimizing $f_{t+1} (\alpha \w_{t})$ over $\alpha$ gives the optimal scaling $\alpha_{t}^* = \frac{\w_{t}^\top \widehat{\B} \w_{t}}{\w_{t}^\top \widehat{\A}_{\lambda} \w_{t}}$ (and this quantity is also invariant to the length of $\w_{t}$). Observe that $\alpha_{t}^*$ naturally measures the quality of $\w_t$: As $\w_t$ converges to $\widehat{\w}$, $\alpha_{t}^*$ converges to $\beta_1$. This initialization technique plays an important role in showing the linear convergence of our algorithm, and was used by~\citet{Ge_16a} for their standard power iterations (alternating least squares) scheme for CCA. 

\begin{proof}[Proof of Lemma~\ref{lem:warm-start}]
 With the given initialization, we have
 \begin{align*}
   f_{t+1} (\alpha_t^* \w_t) - f_{t+1}^* & \le 
   f_{t+1} (\beta_1 \w_t) - f_{t+1}^* \\
   & = \frac{\beta_1^2 \rr_{t}^\top \widehat{\M}_{\lambda}^{-1} \rr_{t}}{2} - \beta_1 \rr_{t}^\top \rr_{t} + \frac{\rr_{t} \widehat{\M}_{\lambda} \rr_{t}}{2} \\
   & = \frac{\norm{\rr_{t}}^2}{2} \sum_{i=1}^d  \xi_{ti}^2 \left( \frac{\beta_1^2}{\beta_i} - 2 \beta_1 + \beta_i \right) \\
   & = \frac{\norm{\rr_{t}}^2}{2} \sum_{i=1}^d  \frac{\xi_{ti}^2}{\beta_i} \left( \beta_1 - \beta_i \right)^2 \\
   & \le \frac{(\w_t^\top \widehat{\B} \w_t)}{2} \cdot \beta_1^2 \sum_{i=2}^d  \frac{\xi_{ti}^2}{\beta_i} .
 \end{align*} 
 Therefore, in view of~\eqref{e:epsilon-t}, it suffices to set the ratio between the initial and the final error of $f_{t+1}$ to
 \begin{align*}
   \max \left( 1, G (\rr_{t}) \right) \cdot \frac{16 \beta_1^2}{\left( \beta_1 - \beta_2 \right)^2}.
 \end{align*}
 In the initial phase, $G (\rr_{t})$ is large, we can set the ratio to be $G (\rr_{0}) \cdot \frac{16 \beta_1^2}{\left( \beta_1 - \beta_2 \right)^2}$, until it is reduced to $1$ after  $\calO \left(\log G (\rr_{0}) \right)$ iterations. Afterwards, we can set the ratio to be the constant of $\frac{16 \beta_1^2}{\left( \beta_1 - \beta_2 \right)^2}$, until we reach the desired accuracy. Observe that
 \begin{align*}
   \frac{\beta_1^2}{\left( \beta_1 - \beta_2 \right)^2} = \left(
     \frac{\frac{1}{\lambda-\widehat{\rho}_1}}{\frac{1}{\lambda-\widehat{\rho}_1}-\frac{1}{\lambda-\widehat{\rho}_2}}
   \right)^2 
   = \left(
     \frac{\lambda-\widehat{\rho}_2}{\widehat{\rho}_1 - \widehat{\rho}_2}
   \right)^2  
   \le (u+1)^2 \le 4.
 \end{align*}
\end{proof}

\subsection{Time complexity of SVRG for finite sum with nonconvex component}
\label{sec:lsq-finite-sum}

\begin{lemma}[Time complexity of SVRG for~\eqref{e:lsq-finite-sum}]
  With the initialization $\alpha_{t}^* \w_t$, SVRG outputs an $\w_{t+1}$ such that $f_{t+1} (\w_{t+1}) - f_{t+1}^* \le \epsilon_t (\w_t^\top \widehat{\B} \w_t)$ in time
\begin{align*}
  \calO \left( d (N + \kappa^2) \log \left( 64 \max\left( G(\rr_t),
  1\right) \right) \cdot \kappa \right), 
\end{align*}
  where $\kappa = \frac{\max_i \, L_i}{\Lambda}$ with $L_i$ being the gradient Lipschitz constant of $f_{t+1}^i$, and $\Lambda$ is the strongly-convex constant of $f_{t+1}$. 
  Futhermore, if we sample each component $f_{t+1}^i$ non-uniformly with probability proportional to $L_i^2$ for the SVRG stochastic updates, we have instead $\kappa=\sqrt{\frac{\frac{1}{N} \sum_{i=1}^N \, L_i^2}{\Lambda^2}}$.
\end{lemma}
Although not explicitly stated by~\citet{GarberHazan15c}, the result for
non-uniform sampling is straightforward by a careful investigation of
their analysis; we provide detailed proof of this result in Appendix~\ref{sec:nonuniform-svrg}. 
The effect of improved dependence on $L_i$'s through non-uniform sampling agrees with related work~\citep{XiaoZhang14a}. The purpose of the non-uniform sampling variant is to bound $\kappa^2$ with high probability for sub-Gaussian/regular polynomial-tail inputs.

\subsection{Bounding the condition number for SVRG}
\label{sec:bounding-kappa-erm}
The next lemma upper-bounds the ``condition number'' $\kappa^2$.

\begin{lemma} \label{lem:bounding-kappa-erm}
  Solving $\min\limits_{\w}\, f_{t+1} (\w)$ using SVRG with non-uniform sampling, we have 
  $\kappa^2 = \calO \left( \frac{d^2}{\widehat{\Delta}^2 \gamma^2} \right)$ for the sub-Gaussian/regular polynomial-tail classes 
with high probability over the sample set, 
  and $\kappa^2 = \calO \left( \frac{1}{\widehat{\Delta}^2 \gamma^2} \right)$ for the bounded class. 
\end{lemma}
\begin{proof}
  The gradient Lipschitz constant $L_i$ is bounded by the largest eigenvalue (in absolute value) of its Hessian
  \begin{align*}
    \Q_{\lambda}^i =
    \left[
      \begin{array}{cc}
        \lambda  \x_i \x_i^\top  & - \x_i \y_i^\top \\
        - \y_i \x_i^\top & \lambda \y_i \y_i^\top
      \end{array}
    \right],
  \end{align*}
  and the largest eigenvalue is defined as
  \begin{align*}
    \max_{\g_x \in \bbR^{d_x},\g_y \in \bbR^{d_y}} \ \beta := \abs{ [\g_x^\top,\g_y^\top] \Q_{\lambda}^i \left[ \begin{array}{c} \g_x \\ \g_y \end{array} \right] } 
    \qquad \text{s.t.} \quad \norm{\g_x}^2 + \norm{\g_y}^2 = 1.
  \end{align*}
  We have
  \begin{align*}
    \beta & = \abs{ \lambda (\g_x^\top \x_i)^2 + \lambda (\g_y^\top \y_i)^2 - 2 (\g_x^\top \x_i) (\g_y^\top \y_i) } \\
    & \le \lambda (\g_x^\top \x_i)^2 + \lambda (\g_y^\top \y_i)^2 + 2 \abs{\g_x^\top \x_i} \abs{\g_y^\top \y_i} \\
    & \le  \lambda (\g_x^\top \x_i)^2 + \lambda (\g_y^\top \y_i)^2 + (\g_x^\top \x_i)^2 + (\g_y^\top \y_i)^2 \\
    & = (\lambda + 1) \left( (\g_x^\top \x_i)^2 + (\g_y^\top \y_i) \right) \\
    & \le (\lambda + 1) \left( \norm{\g_x}^2 \norm{\x_i}^2 + \norm{\g_y}^2 \norm{\y_i}^2) \right) \\
    & \le (\lambda + 1) \cdot \max \left(\norm{\x_i}^2,  \norm{\y_i}^2 \right) \\
    & \le (\lambda + 1) \cdot \left( \norm{\x_i}^2 + \norm{\y_i}^2  \right)
  \end{align*}
  where we have used the Cauchy-Schwarz inequality in the third inequality.
  
  Note that, for bounded inputs, we have $\norm{\x_i}^2 + \norm{\y_i}^2 \le 2$ and so $L_i^2 \le 4 (\lambda+1)^2$ for all $i=1,\dots, N$. For sub-Gaussian/regular polynomial-tail inputs, we have 
\begin{align*}
\frac{1}{N} \sum_{i=1}^N L_i^2 \le (\lambda+1)^2 \cdot \frac{1}{N} \sum_{i=1}^N \left( \norm{\x_i}^2 + \norm{\y_i}^2 \right)^2 = \calO ((\lambda+1)^2 d^2)
\end{align*}
with high probability in view of Remark~\ref{rmk:data-norm-bound}. 
  
  On the other hand, we have shown that $\Lambda = \sigma_{\min} \left( \A_{\lambda}  \right) \ge
  (\lambda-\widehat{\rho}_1) \gamma / 2$. Recalling
  $\lambda=\widehat{\rho}_1 + c \widehat{\Delta}$ with $c \in (0,1)$, we
  have $\lambda \le 2$ and $\Lambda \ge c \widehat{\Delta} \gamma / 2$. Combining this with the data norm bound above yields the desired result. 
\end{proof}

\subsection{Proof of Theorem~\ref{thm:total-time-stochastic-any-N}}

\begin{proof} Since $\frac{ {\uu_T}^\top \Sxx \widehat{\uu}}{\norm{\Sxx^{\frac{1}{2}} {\uu_T}}} \le 1$ and $\frac{ {\vv_T}^\top \Syy \widehat{\vv}}{\norm{\Syy^{\frac{1}{2}} {\vv_T}}} \le 1$, it suffices to require 
  \begin{align*}
    \frac{ {\uu_T}^\top \Sxx \widehat{\uu}}{\norm{\Sxx^{\frac{1}{2}} {\uu_T}}} + \frac{ {\vv_T}^\top \Syy \widehat{\vv}}{\norm{\Syy^{\frac{1}{2}} {\vv_T}}} \ge 2 - \eta.
  \end{align*}

  According to Lemma~\ref{lem:alignment-transfer} (making the identification that $\aa = \Sxx^{\frac{1}{2}} \widehat{\uu}$, $\x = \Sxx^{\frac{1}{2}} \uu_T$, $\b = \Syy^{\frac{1}{2}} \widehat{\vv}$, and $\y = \Syy^{\frac{1}{2}} \vv_T$), it then suffices to have
  \begin{align}
    \cos \theta_T = 
    \frac{1}{\sqrt{2}} 
    \frac{{ \widehat{\uu}^\top \Sxx \uu_T + \widehat{\vv}^\top \Syy \vv_T }}
    {\sqrt{{\uu}_T^\top \Sxx {\uu}_T + {\vv}_T ^\top \Syy {\vv}_T}}
    \ge 1 - \frac{\eta}{8}.
  \end{align}
  Since $\cos \theta_T = \sqrt{1 - \sin^2 \theta_T} \ge 1 - \sin^2
  \theta_T$, we just need $\abs{\sin \theta_T} \le
  \frac{\sqrt{\eta}}{\sqrt{8}}$, and we ensure it by requiring
  $G(\rr_T)\le \frac{\sqrt{\eta}}{\sqrt{8}}$.
 Applying results from the previous sections, we need to solve $\calO
  \left( \log  \frac{G (\rr_0)}{\eta} \right)$ linear systems, and the time
  complexity for solving each is at most $\calO\left( (N + \kappa^2) \log
    \rbr{G(\rr_0) \cdot \kappa} \right)$ for SVRG. 

It remains to bound $G(\rr_0)$. By the definition of $G(\cdot)$, we have
\begin{align*}
G(\rr_0) =  \frac{ \sqrt{ \sum_{i=2}^d  \xi_{0i}^2 / \beta_i }}{\sqrt{\xi_{01}^2 / \beta_1}}
\le \sqrt{\frac{\beta_1}{\beta_d}} \cdot \frac{1}{\abs{\xi_{01}}}
\end{align*}
where $\abs{\xi_{01}}$ is the alignment between $\rr_0$ and $\p_1$ which, 
by the relationship between $\rr_0$ and $\w_0$, satisfy
\begin{align*}
\abs{\rr_0 ^\top \p_1} & = \frac{ \abs{\tw_0^\top \widehat{\B}^{\frac{1}{2}} \p_1} }
{\norm{\widehat{\B}^{\frac{1}{2}} \tw_0}}
\ge \frac{\abs{\tw_0^\top (\widehat{\B}^\frac{1}{2}
  \p_1)}}{\sigma_{\max}(\widehat{\B}^\frac{1}{2} ) \cdot \norm{\tw_0}}
= \frac{\norm{\widehat{\B}^\frac{1}{2} \p_1} \cdot \abs{(\tw_0/\norm{\tw_0})^\top (\widehat{\B}^\frac{1}{2} \p_1 /
  \norm{\widehat{\B}^\frac{1}{2} \p_1})}}
{\sigma_{\max}(\widehat{\B}^\frac{1}{2} )} \\
& \ge \frac{\sigma_{\min}(\widehat{\B}^\frac{1}{2})}{\sigma_{\max}(\widehat{\B}^\frac{1}{2})}
\cdot \abs{ \rbr{\frac{\tw_0}{\norm{\tw_0}}}^\top
  \rbr{\frac{\widehat{\B}^\frac{1}{2} \p_1}{\norm{\widehat{\B}^\frac{1}{2}\p_1}}} }.
\end{align*}
According to the way $\tw_0$ is initialized
and~\citet[Lemma~5]{Arora_09a}, we have with probability at least $1-C$ that
$\abs{ \rbr{\frac{\tw_0}{\norm{\tw_0}}}^\top
  \rbr{\frac{\widehat{\B}^\frac{1}{2}
      \p_1}{\norm{\widehat{\B}^\frac{1}{2}\p_1}}} } \ge
\frac{C}{\sqrt{d}}$.
On the other hand, we have $\frac{\beta_1}{\beta_d} = \calO (\frac{1}{\widehat{\Delta}})$
and
$\frac{\sigma_{\max}(\widehat{\B}^\frac{1}{2})}{\sigma_{\min}(\widehat{\B}^\frac{1}{2})}
= \calO \rbr{\frac{1}{\sqrt{\gamma}}}$. 
Combining these results yields that 
$G(\rr_0) = \calO \rbr{\sqrt{\frac{d}{\widehat{\Delta} \gamma}}}$ with high probability. 
Then the theorem follows.
\end{proof}

\subsection{Proof of Corollary~\ref{cor:total-time-erm-stochastic}}

\begin{proof}
  Denote $\widetilde{\rr} := \frac{1}{\sqrt{2}} \left[ \begin{array}{c} \Sxx^{\frac{1}{2}} \uu_T / \norm{\Sxx^{\frac{1}{2}} \uu_T}\\ \Syy^{\frac{1}{2}} \vv_T / \norm{\Syy^{\frac{1}{2}} \vv_T} \end{array}\right]$, with $\norm{\widetilde{\rr}}=1$. Assume without loss of generality that $\norm{\Sxx^{\frac{1}{2}} \uu_T}=\norm{\Syy^{\frac{1}{2}} \vv_T}=1$; this does not affect our measure of alignment, and can be ensured by a final (separate) normalization step with cost $\calO (Nd)$~\citep{Wang_16b}.

Apply Lemma~\ref{lem:approx-error-erm} with
$\epsilon^\prime=\frac{\sqrt{\epsilon} \Delta}{8}$; requiring that
$\epsilon^\prime = \frac{\sqrt{\epsilon} \Delta}{8} \le C_d \gamma^2$ 
as in Corollary~\ref{cor:sample-complexity-canoncorr} leads
to the extra condition that $\epsilon \le \frac{64 C_d^2 \gamma^4}{\Delta^2}$.

  With the specified sample complexity, we have that with high probability 
  \begin{align} \label{e:emp-gap}
    \norm{ \T - \widehat{\T} } \le \frac{\sqrt{\epsilon} \Delta}{8} \le \frac{\Delta}{8}.
  \end{align}
  In view of the Weyl's inequality, \eqref{e:emp-gap} implies that $\widehat{\Delta} \ge \frac{3 \Delta}{4}$.

  Let $\rr^* = \frac{1}{\sqrt{2}} \left[ \begin{array}{c} \Exx^{\frac{1}{2}} \uu^* \\ \Eyy^{\frac{1}{2}} \vv^* \end{array}\right]$ be the top eigenvector of $\C$. And recall $\widehat{\rr} := \frac{1}{\sqrt{2}} \left[ \begin{array}{c} \Sxx^{\frac{1}{2}} \widehat{\uu}  \\ \Syy^{\frac{1}{2}} \widehat{\vv} \end{array}\right]$ is the top eigenvector of $\widehat{\C}$.  
  According to the Davis-Kahan $\sin \theta$ theorem~\citep{DavisKahan70a}, with the number of samples given in the theorem, the top eigenvectors of $\C$ and $\widehat{\C}$ are well aligned:
  \begin{align*} 
    \sin^2 \theta \le \frac{\norm{\C - \widehat{\C}}^2}{\Delta^2} \le \frac{\epsilon}{64}
  \end{align*}
  where $\theta$ is the angle between $\rr^*$  and $\widehat{\rr}$. This implies that 
\begin{align*}
\widehat{\rr}^\top {\rr^*} = \cos \theta = \sqrt{1 - \sin^2 \theta} \ge 1 - \sin^2 \theta \ge 1 - \frac{\epsilon}{64}. 
\end{align*}

  We now show that the theorem follows if we manage to solve the ERM objective so accurately that
  \begin{align} \label{e:necessary-alignment-erm}
    {\widetilde{\rr}^\top \widehat{\rr}} = 
    \frac{1}{2} \left( \frac{\widehat{\uu}^\top \Sxx \uu_T}{\sqrt{\uu_T^\top \Sxx \uu_T}} + \frac{\widehat{\vv}^\top \Syy \vv_T}{\sqrt{\vv_T^\top \Syy \vv_T}} \right)
    \ge 1 - \frac{\epsilon^2}{8192}.
  \end{align}
  To see this, first observe that~\eqref{e:necessary-alignment-erm} implies
  \begin{align*}
\norm{\widetilde{\rr} - \widehat{\rr}} = \sqrt{ 2 - 2 (\widetilde{\rr}^\top \widehat{\rr}) } \le \frac{\epsilon}{64} ,
  \end{align*}
  and as a result
\begin{align*}
\widetilde{\rr}^\top \rr^* \ge \widehat{\rr}^\top \rr^* - \abs{ (\widetilde{\rr} - \widehat{\rr})^\top \rr^*} 
\ge \widehat{\rr}^\top \rr^* - \norm{\widetilde{\rr} - \widehat{\rr}} \ge 1 - \frac{\epsilon}{32}. 
\end{align*}
Consequently, we have
\begin{align*}
\frac{1}{2} \left( \norm {\Exx^{\frac{1}{2}} \uu^* - \Sxx^{\frac{1}{2}} \uu_T}^2 + \norm {\Eyy^{\frac{1}{2}} \vv^* - \Syy^{\frac{1}{2}} \vv_T}^2 \right) 
= \norm{\widetilde{\rr} -  \rr^*}^2 = 2 \left( 1 - \widetilde{\rr}^\top \rr^* \right) \le \frac{\epsilon}{16} 
\end{align*}
and so $\max \left( \norm{ \Exx^{\frac{1}{2}} \uu^* - \Sxx^{\frac{1}{2}} \uu_T }^2,\, \norm{ \Eyy^{\frac{1}{2}} \vv^* - \Syy^{\frac{1}{2}} \vv_T }^2 \right) \le \frac{\epsilon}{8}$.  
We are now in the same situation as~\eqref{e:erm-fix-normalization}; we can fix the incorrect normalization of $\widetilde{\rr}$ analogously and then our lemma follows. 

  It remains to show the time complexity to achieve~\eqref{e:necessary-alignment-erm}. According to Lemma~\ref{lem:alignment-transfer}, it suffices to have
  \begin{align}
    \cos \theta_T = 
    \frac{1}{\sqrt{2}} 
    \frac{{ \widehat{\uu}^\top \Sxx \uu_T + \widehat{\vv}^\top \Syy \vv_T }}
    {\sqrt{{\uu}_T^\top \Sxx {\uu}_T + {\vv}_T ^\top \Syy {\vv}_T}}
    \ge 1 - \frac{\epsilon^2}{2^{15}}.
  \end{align}
  In turn, it suffices to have $\abs{\sin \theta_T} \le
  \frac{\epsilon}{256}$ and we ensure it by requiring $G(\rr_T)\le
  \frac{\epsilon}{256}$. We obtain the stated time complexity by applying
  Theorem~\ref{thm:total-time-stochastic-any-N} with $\eta = \frac{\epsilon}{256}$.


\end{proof}

\section{Proofs for Section~\ref{sec:alg-shift-and-invert-online}}
\label{sec:append-streaming}

\subsection{Proof of Lemma~\ref{lem:online-transfer-align}}
\begin{proof}
  The desired result is a direct consequence of~Lemma~\ref{lem:alignment-transfer}, by making the identification that 
  \begin{align*}
    \aa = \Exx^{\frac{1}{2}} \uu^*, \qquad \x = \Exx^{\frac{1}{2}} \uu_T, \qquad
    \b = \Eyy^{\frac{1}{2}} \vv^*, \qquad \y = \Eyy^{\frac{1}{2}} \vv_T. 
  \end{align*}
\end{proof}

\subsection{Parameters of Streaming SVRG for stochastic least squares}
\label{sec:streaming-parameters}


We divide Lemma~\ref{lem:streaming-parameters}  into the following three propositions.

 \begin{proposition}[Strong convexity] \label{lem:streaming-strong-convexity}
   For any $\w,\w^\prime \in \bbR^d$, we have
   \begin{align*}
     f_{t+1} (\w) \ge f_{t+1} (\w^\prime) + \left<\nabla f_{t+1} (\w^\prime), \w - \w^\prime \right> + \frac{\mu}{2} \norm{\w - \w^\prime}^2
   \end{align*}
   where $\mu:= \frac{\gamma}{\beta_1} \ge C \Delta \gamma$ for some $C>0$.
 \end{proposition}
\begin{proof}
  Just observe that the Hessian of $f_{t+1} (\w)$ is $\A_{\lambda}={\B}^{\frac{1}{2}} {\M}_{\lambda}^{-1} {\B}^{\frac{1}{2}}$, whose eigenvalues are bounded from below:  $\sigma_{\min} \left(\A_{\lambda}\right) \ge (\lambda-\rho_1) \cdot \sigma_{\min} \left( \B \right) =\gamma / \beta_1$. The lemma follows from the assumption that $\lambda=\rho_1 + c \Delta$ for $c \in (0,1)$.
\end{proof}

 \begin{proposition}[Streaming smoothness]\label{lem:streaming-smoothness}
   For any $\w \in \bbR^d$, we have
   \begin{align*}
     \bbE \left[ 
       \norm{ \nabla \phi_{t+1} (\w) - \nabla \phi_{t+1} (\w_{t+1}^*) }^2 \right] 
     \le 2 S  \left( f_{t+1} (\w) - f_{t+1}^* \right)
   \end{align*}
   where $S = \calO \left( \frac{d \beta_1}{\gamma} \right) = \calO \left( \frac{d}{\Delta \gamma} \right)$ for the sub-Gaussian/regular polynomial-tail classes, and $S = \calO \left( \frac{\beta_1}{\gamma} \right) = \calO \left( \frac{1}{\Delta \gamma} \right)$ for the bounded class.
 \end{proposition} 
\begin{proof} Observe that
  \begin{align*}
    \nabla \phi_{t+1} (\w) = 
    \left[
      \begin{array}{cc}
        \lambda  \x \x^\top  & - \x \y^\top \\
        - \y \x^\top & \lambda \y \y^\top
      \end{array}
    \right]
    \w
    - 
    \left[
      \begin{array}{cc}
        \x \x^\top  & \\
        & \y \y^\top
      \end{array}
    \right] \w_t .
  \end{align*}
  As shown in Lemma~\ref{lem:bounding-kappa-erm}, this gradient function is Lipschitz continuous:
  \begin{align*}
    \norm{ \nabla \phi_{t+1} (\w) - \nabla \phi_{t+1} (\w_{t+1}^*) }
    \le (\lambda+1) \cdot \sup \left(\norm{\x},\, \norm{\y}\right) \cdot \norm{ \w - \w_{t+1}^*}. 
  \end{align*}
  Note that $\lambda \le \rho_1 + u \Delta$ where $\rho_1 \le 1$, $\Delta \le 1$, and $u < 1$ by assumption, and thus $\lambda\le 2$. As a result, we obtain
  \begin{align*}
    \bbE \norm{ \nabla \phi_{t+1} (\w) - \nabla \phi_{t+1} (\w_{t+1}^*) }^2
    \le 9 \bbE \left[ \norm{\x}^2 + \norm{\y}^2 \right] \cdot \norm{ \w - \w_{t+1}^*}^2.
  \end{align*}

For the distributions of $P(\x,\y)$ considered here, $\bbE \norm{\x}^2$ and $\bbE \norm{\y}^2$ are both $\calO (d)$ for the sub-Gaussian/regular polynomial-tail inputs (see Remark~\ref{rmk:data-norm-bound}), and bounded by $1$ for the bounded inputs.

  On the other hand, according to Lemma~\ref{lem:streaming-strong-convexity}, we have
  \begin{align*}
    f (\w) - f (\w_{t+1}^*) \ge C \Delta \gamma \norm{\w - \w_{t+1}^*}^2 
  \end{align*}
for some $C>0$.

  Combining the above two inequalities gives the desired result.
\end{proof}


 \begin{proposition}[Streaming variance]\label{lem:streaming-variance}
   We have 
   \begin{align*}
     \bbE \left[ \frac{1}{2} \norm{\nabla \phi (\w_{t+1}^*)}^2_{\left( \nabla^2 f(\w_{t+1}^*)\right)^{-1}} \right] \le \sigma^2.
   \end{align*}
   where $\sigma^2= \calO \left( d \beta_1^3 \norm{\rr_{t}}^2 \right)$ for the sub-Gaussian/regular polynomial-tail classes, and $\sigma^2= \calO \left( \frac{\beta_1^3 \norm{\rr_{t}}^2}{\gamma^2} \right)$ for the bounded class.
 \end{proposition} 
\begin{proof} Observe that $\w_{t+1}^* = \A_{\lambda}^{-1} \B \w_{t} $ and 
  \begin{align*}
    \nabla \phi (\w_{t+1}^*) =  \left( \left[
        \begin{array}{cc}
          \lambda  \x \x^\top  & - \x \y^\top \\
          - \y \x^\top & \lambda \y \y^\top
        \end{array}
      \right] \A_{\lambda}^{-1} \B 
      - 
      \left[
        \begin{array}{cc}
          \x \x^\top  & \\
          & \y \y^\top
        \end{array}  
      \right] \right) \w_{t} .
  \end{align*}
  Define the shorthands $\D =  \B^{-\frac{1}{2}}
  \left[
    \begin{array}{cc}
      \lambda  \x \x^\top  & - \x \y^\top \\
      - \y \x^\top & \lambda \y \y^\top
    \end{array}
  \right] \B^{-\frac{1}{2}}$
  and 
  $\E = \B^{-\frac{1}{2}}
  \left[
    \begin{array}{cc}
      \x \x^\top  & \\
      & \y \y^\top
    \end{array}
  \right] \B^{-\frac{1}{2}}$.

  Then we have 
  \begin{align}
    \bbE \left[ \frac{1}{2} \norm{\nabla \phi (\w_{t+1}^*)}^2_{\left( \nabla^2 f(\w_{t+1}^*)\right)^{-1}} \right] 
    & = \bbE \left[ \frac{1}{2} \norm{\nabla \phi (\w_{t+1}^*)}^2_{\A_{\lambda}^{-1}} \right]  \nonumber \\
    & = \bbE \left[ \frac{1}{2} \norm{ \B^{-\frac{1}{2}} \nabla \phi (\w_{t+1}^*)}^2_{\B^{\frac{1}{2}} \A_{\lambda}^{-1} \B^{\frac{1}{2}}} \right] \nonumber  \\
    & = \frac{1}{2} \bbE \left[ \left(\B^{\frac{1}{2}} \w_t\right)^\top    (\M_{\lambda} \D  -  \E) \cdot \M_{\lambda} \cdot (\D \M_{\lambda} - \E )  \left(\B^{\frac{1}{2}} \w_t\right) \right]  \nonumber  \\  \label{e:streaming-variance-1}
    & = \frac{1}{2} \bbE \left[ \rr_t^\top    (\M_{\lambda} \D  -  \E) \cdot \M_{\lambda} \cdot (\D \M_{\lambda} - \E )  \rr_t \right] .
  \end{align}

  \textbf{Bounded case\ } For the bounded case where $\sup \left( \norm{\x}^2, \norm{\y}^2 \right) \le 1$, the derivation is relatively simple. We can bound $\norm{\D}\le \frac{3}{\gamma}$ and $\norm{\E} \le \frac{1}{\gamma}$, and thus 
  \begin{align*}
    \bbE \left[ \frac{1}{2} \norm{\nabla \phi (\w_{t+1}^*)}^2_{\left( \nabla^2 f(\w_{t+1}^*)\right)^{-1}} \right] 
    & \le \frac{\norm{\M_{\lambda}} \norm{\rr_t}^2 }{2} \bbE \norm{\D \M_{\lambda} - \E}^2  \\
    & \le \beta_1  \norm{ \rr_t}^2  \left( \bbE \norm{\D \M_{\lambda}}^2 + \bbE \norm{\E}^2 \right) \\
    & = \calO \left( \frac{\beta_1^3 \norm{\rr_t}^2}{\gamma^2} \right)
  \end{align*}
  where we have used the triangle inequality and the fact that $(x+y)^2 \le 2 x^2 + 2 y^2$ in the second inequality. 

  \textbf{Sub-Gaussian/regular polynomial tail cases\ } 
  We now omit the subscript $\lambda$ in $\M_{\lambda}$ and $t$ from iterates for convenience. Using the fact that $\norm{\x + \y}^2 \le 2 \norm{\x}^2 + 2  \norm{\y}^2$ with $\x = \M^{\frac{1}{2}} \D \M \rr$ and $\y=\M^{\frac{1}{2}} \E \rr$, we continue from~\eqref{e:streaming-variance-1} and obtain
  \begin{align*}
    \bbE \left[ \frac{1}{2} \norm{\nabla \phi (\w_{t+1}^*)}^2_{\left( \nabla^2 f(\w_{t+1}^*)\right)^{-1}} \right] 
    \le \bbE \left[ \rr^\top \M \D \M \D \M \rr \right] + \bbE \left[\rr^\top \E \M \E \rr \right].
  \end{align*}
  Introduce the notation $\uu=\Exx^{-\frac{1}{2}} \x$, $\vv=\Eyy^{-\frac{1}{2}} \y$, and parition $\rr$ and $\M$ according to $(\x, \y)$:
  \begin{align*}
    \rr=\begin{bmatrix}
      \rr_x \\
      \rr_y
    \end{bmatrix},\qquad \M=\begin{bmatrix}
      \M_{xx} & \M_{xy} \\
      \M_{yx} & \M_{yy}
    \end{bmatrix}.
  \end{align*}
  In view of Lemma~\ref{lem:sub-gaussian-moments}, we can assume $\max \left( \bbE \norm{\uu}^4,\, \bbE \norm{\vv}^4 \right) \le C d^2$. From now on, we use $C$ for a generic constant whose specific value may change between appearances.

  We have 
  \begin{align*}
    \E \M \E=\begin{bmatrix}
      \uu\uu^\top  \M_{xx} \uu\uu^\top  & \uu\uu^\top  \M_{xy} \vv\vv^\top  \\
      \vv\vv^\top  \M_{yx} \uu\uu^\top  & \vv\vv^\top  \M_{yy} \vv\vv^\top 
    \end{bmatrix}.
  \end{align*}
  and thus
  \begin{align*}
    \rr^\top \E \M \E \rr= \rr_x^\top \uu\uu^\top \M_{xx}\uu\uu^\top \rr_x + \rr_y^\top \vv\vv^\top \M_{yy} \vv\vv^\top \rr_y + 2\rr_x^\top \uu\uu^\top \M_{xy}\vv\vv^\top \rr_y.
  \end{align*}
  We have for the first term that 
  \begin{align*}
    \bbE \left[ \rr_x^\top \uu \uu^\top \M_{xx} \uu \uu^\top \rr_x \right] 
    & = \bbE \left[ \abs{\rr_x^\top \uu}^2 \uu^\top \M_{xx} \uu \right] \\
    &\leq \sqrt{\bbE \abs{\rr_x^\top \uu}^4 } \sqrt{\bbE \abs{\uu^\top \M_{xx} \uu}^2} \\
    &\leq C \norm{\rr_x}^2 \sqrt{\bbE \abs{\uu^\top \M_{xx} \uu}^2} \\
    &\leq C \norm{\M_{xx}} \norm{\rr_x}^2 \sqrt{\bbE \norm{\uu}^4 } \\
    &\leq C \norm{\M} \norm{\rr_x}^2 d.
  \end{align*}
  Similar arguments also lead to
  \begin{align*}
    \bbE \left[ \rr_y^\top \vv\vv^\top \M_{yy} \vv\vv^\top \rr_y \right] \leq C \norm{\M} \norm{\rr_y}^2 d.
  \end{align*}
  For the third term, we have
  \begin{align*}
    \bbE \left[ \rr_x^\top \uu\uu^\top \M_{xy}\vv\vv^\top \rr_y \right] 
    & \leq \sqrt{\bbE \abs{\rr_x^\top \uu}^2 \abs{\rr_y^\top \vv}^2} \sqrt{\bbE \abs{\uu^\top \M_{xy} \vv}^2} \\
    &\leq  \norm{\M} \left(\bbE \abs{\rr_x^\top \uu}^4 \right)^{\frac{1}{4}} \left(\bbE \abs{\rr_y^\top \vv}^4 \right)^{\frac{1}{4}} \left(\bbE \norm{\uu}^4 \right)^{\frac{1}{4}} \left(\bbE \norm{\vv}^4 \right)^{\frac{1}{4}} \\
    &\leq C \norm{\M} \norm{\rr_x} \norm{\rr_y} d.
  \end{align*}
  Therefore,
  \begin{align*}
    \bbE \left[\rr^\top \E \M \E \rr \right]\leq C \norm{\M} \norm{\rr}^2 d.
  \end{align*}
  Now we need to bound $\bbE \left[ \rr^\top \M \D \M \D \M \rr \right]$. Using the fact that $\norm{\x + \y}^2 \le 2 \norm{\x}^2 + 2  \norm{\y}^2$ with $\x = \M^{\frac{1}{2}} \D_1 \M \rr$ and $\y=\M^{\frac{1}{2}} \D_2 \M \rr$, this can be bounded by two terms:
  \begin{align*}
    \bbE \left[\rr^\top \M \D \M \D \M \rr \right] \leq 2 \bbE \left[\rr^\top \M \D_1 \M \D_1 \M \rr \right] + 2 \bbE \left[\rr^\top \M \D_2 \M \D_2 \M \rr\right]
  \end{align*}
  where
  \begin{align*}
    \D_1=\lambda\begin{bmatrix}
      \uu\uu^\top  & 0 \\
      0 & \vv\vv^\top 
    \end{bmatrix},\qquad 
    \D_2=-\begin{bmatrix}
      0 & \uu \vv^\top  \\
      \vv\uu^\top  & 0
    \end{bmatrix}.
  \end{align*}
 The bound for $\bbE \left[\rr^\top \M \D_1 \M \D_1 \M \rr\right]$ can be derived using the same argument that bounds $\bbE\left[\rr^\top \E \M \E \rr \right]$ (now $\M \rr$ plays the role of $\rr$ in bounding $\bbE\left[\rr^\top \E \M \E \rr \right]$), and thus we have
 \begin{align*}
   \bbE\left[ \rr^\top \M \D_1 \M \D_1 \M \rr \right] \leq C \lambda^2 \norm{\M} \norm{\M \rr}^2 d \leq C  \norm{\M}^3 \norm{\rr}^2 \lambda^2 d.
\end{align*}
 Finally, we bound $\bbE\left[\rr^\top \M \D_2 \M \D_2 \M \rr \right]$. Note that
 \begin{align*}
   -\D_2 \M \D_2=\begin{bmatrix}
   \uu \vv^\top \M_{yy} \vv\uu^\top  & \uu \vv^\top \M_{yx}\uu \vv \\
   \vv \uu^\top \M_{xy} \vv \uu^\top  & \vv \uu^\top \M_{xx}\uu \vv^\top 
 \end{bmatrix}.
\end{align*}
 Let
\begin{align*}
  \M \rr = \begin{bmatrix}
   \m_x \\
   \m_y
 \end{bmatrix},
\end{align*}
 and then
 \begin{align*}
   - \rr^\top \M \D_2 \M \D_2 \M \rr=\m_x^\top \uu \vv^\top \M_{yy} \vv \uu^\top \m_x + \m_y^\top \vv\uu^\top \M_{xx}\uu \vv^\top \m_y + 2 \m_x^\top \uu \vv^\top \M_{yx}\uu \vv^\top \m_y.
\end{align*}
 Similarly to what we have done above,
 \begin{align*}
   \bbE \abs{\m_x^\top \uu \vv^\top \M_{yy} \vv \uu^\top \m_x} &\leq \sqrt{\bbE \abs{\m_x^\top \uu}^4} \sqrt{\bbE \abs{\vv^\top \M_{yy} \vv}^2} \\
   &\leq C \norm{\M} \norm{\m_x}^2 d \\
   &\leq C \norm{\M}^3 \norm{\rr}^2 d.
 \end{align*}
 The same bound also holds for $\bbE \abs{\m_y^\top \vv\uu^\top \M_{xx}\uu \vv^\top \m_y}$ with the same argument. For the term $\bbE \abs{\m_x^\top \uu \vv^\top \M_{yx} \uu \vv^\top \m_y}$, we have
\begin{align*}
  \bbE \abs{\m_x^\top \uu \vv^\top \M_{yx} \uu \vv^\top \m_y} &\leq  \norm{\M} \left(\bbE \abs{\m_x \uu}^4 \right)^{\frac{1}{4}}\left(\bbE \abs{\m_y \vv}^4 \right)^{\frac{1}{4}}\left(\bbE \norm{\uu}^4 \right)^{\frac{1}{4}}\left(\bbE \norm{\vv}^4 \right)^{\frac{1}{4}} \\
  &\leq C \norm{\M} \norm{\m_x} \norm{\m_y} d \\
  &\leq C \norm{\M}^3 \norm{\rr}^2 d.
\end{align*}
Combining all the terms, and noting that $\lambda \le 2$, we have shown that
\begin{align*}
\bbE \left[\rr^\top \M \D \M \D \M \rr \right] \le C  \norm{\M}^3 \norm{\rr}^2 d.
\end{align*}
And the final bound is
\begin{align*}
\bbE \left[ \rr^\top (\M \D - \E) \M (\D \M - \E) \rr \right] \leq C \left[\norm{\M}^3 + \norm{\M} \right] \norm{\rr}^2 d 
=\calO \left( \beta_1^3 \norm{\rr}^2 d \right).
\end{align*}
\end{proof}

\subsection{Proof of Lemma~\ref{lem:streaming-svrg-sample-complexity-for-f}}

\begin{proof} For notational simplicity, we omit the subscript ${t+1}$ below. 

  According to~\citet[Theorem~4.1]{Frostig_15b}, we have that for iteration $\tau$ of Algorithm~\ref{alg:streaming-svrg}
  \begin{align} 
    \bbE \left[ f (\w^{\tau}) - f^* \right] 
    & \le  \frac{1}{1 - 4 s} \left[
      \left(\frac{S}{\mu m_{\tau} s} + 4 s \right) \bbE \left[ f (\w^{\tau-1}) - f^* \right]  \right. \nonumber \\ \label{e:streaming-one-step}
    & \qquad\qquad\qquad\qquad\qquad \left. + \frac{1 + 2 s}{k_{\tau}} \left(\sqrt{\frac{S}{\mu} \bbE \left[ f (\w^{\tau-1}) - f^* \right] } + \sigma
      \right)^2
    \right].
  \end{align}

  Using the inequality $(x+y)^2 \le 2 (x^2 + y^2)$, it holds that
  \begin{align*}
    \left(\sqrt{\frac{S}{\mu} \bbE \left[ f (\w^{\tau-1}) - f^* \right] } + \sigma \right)^2
    \le \frac{2S}{\mu} \bbE \left[ f (\w^{\tau-1}) - f^* \right] + 2 \sigma^2. 
  \end{align*}

  Now, set for this iteration $s=\frac{c_2}{8}$, $m_{\tau}=\ceil{\frac{S}{\mu c_2^2}}$, and $k_{\tau}=\max\left( \ceil{\frac{S}{\mu c_2}}, \, \ceil{\frac{\sigma^2}{\beta_1 \norm{\rr_{t}}^2 c_3}} \right)$, for some $c_2, c_3 \in (0,1)$. 
  We continue from~\eqref{e:streaming-one-step} and have
  \begin{align*} 
    \bbE \left[ f (\w^{\tau}) - f^* \right] & \le
    \frac{1}{1 - 4 s} \left[
      \left(\frac{S}{\mu m_{\tau} s} + 4 s + \frac{2 + 4 s }{k_{\tau}} \frac{S}{\mu} \right) \bbE \left[ f (\w^{\tau-1}) - f^* \right]
      + \frac{2 + 4 s}{k_{\tau}} \sigma^2 \right] \\
    & \le \frac{1}{1 - c_2/2} \left[ \left(8 c_2 + \frac{c_2}{2} + \frac{2 + 4 c_2}{2} c_2 \right) \bbE \left[ f (\w^{\tau-1}) - f^* \right] + \frac{4 + c_2}{2 k_{\tau}} \sigma^2 \right] \\
    & \le  22 c_2 \cdot \bbE \left[ f (\w^{\tau-1}) - f^* \right]  + 10 c_3 \cdot \frac{\beta_1 \norm{\rr_{t}}^2}{2} .
  \end{align*}
  We can now calculate the number of samples used in this iteration, which is 
  \begin{align} \label{e:sample-one-iteration-AB}
    k_{\tau} + m_{\tau} = \calO \left( \frac{d \beta_1^2}{c_3}  + \frac{d \beta_1^2}{\gamma^2 c_2^2}   \right) 
    = \calO \left( \frac{d}{\Delta^2 c_3}  + \frac{d}{\Delta^2 \gamma^2 c_2^2}  \right) 
  \end{align} 
for sub-Gaussian/regular polynomial-tail inputs, and 
  \begin{align} \label{e:sample-one-iteration-C}
    k_{\tau} + m_{\tau} = \calO \left( \frac{\beta_1^2}{\gamma^2 c_3}  + \frac{\beta_1^2}{\gamma^2 c_2^2}   \right) 
    = \calO \left( \frac{1}{\Delta^2 \gamma^2 c_3} + \frac{1}{\Delta^2 \gamma^2 c_2^2} \right) 
  \end{align} 
for bounded inputs.

  Let us fix $c_2 = \frac{1}{44}$ for $\tau=1,\dots,\Gamma$. In view of our initialization strategy~\eqref{e:streaming-initialization}, setting $c_3=\frac{1}{20}$ for $\tau=1$ gives $ \bbE \left[ f (\w^{1}) - f^* \right] \le  \frac{\beta_1 \norm{\rr_{t}}^2}{2}$. 
  Afterwards, we halve $c_3$ at each outer loop $\tau=2,\dots$, and this makes sure that $\bbE \left[ f (\w^{\tau}) - f^* \right] \le  \frac{\beta_1 \norm{\rr_{t}}^2}{2^{\tau}}$. 

  To achieve the desired accuracy, we need $\Gamma = \calO \left( \log
    \frac{1}{\eta_t} \right)$ outer iterations. Summing~\eqref{e:sample-one-iteration-AB} and~\eqref{e:sample-one-iteration-AB} over $\tau=1,\dots,\Gamma$, and noting $\sum_{\tau=1}^\Gamma 2^{\tau-1} = \calO \left( \frac{1}{\eta_t} \right)$, the total sample complexity is 
  \begin{align*}
    \calO \left(  \frac{d}{\Delta^2} \cdot 20 \sum_{\tau=1}^{\Gamma} 2^{\tau-1} + \frac{44^2 d}{\Delta^2 \gamma^2} \cdot \log  \frac{1}{\eta_t} \right) = \calO \left(  \frac{d}{\Delta^2 \eta_t} + \frac{d}{\Delta^2 \gamma^2} \log \frac{1}{\eta_t} \right)
  \end{align*} 
for sub-Gaussian/regular polynomial-tail inputs, and 
  \begin{align*}
    \calO \left(    \frac{1}{\Delta^2 \gamma^2} \left( 20 \sum_{\tau=1}^{\Gamma} 2^{\tau-1} + 44^2 \cdot \log  \frac{1}{\eta_t} \right) \right) = \calO \left(  \frac{1}{\Delta^2 \gamma^2 \eta_t}  \right)
  \end{align*} 
for bounded inputs (we have dropped the second term since $\log \frac{1}{\eta_t}$ is of lower order compared with $\frac{1}{\eta_t}$).
\end{proof}

\subsection{Proof of Theorem~\ref{thm:sample-complexity-for-online-cca}}

\begin{proof}
  Recall that our streaming CCA algorithm performs shift-and-invert power iterations on the population matrices directly. Following the same argument in the ERM case in Corollary~\ref{cor:total-time-erm-stochastic}, as long as each least squares objective is solved to sufficient accuracy, \ie, 
  \begin{align} \label{e:streaming-svrg-epsilon-t}
    \frac{f_{t+1} (\w_{t+1}) - f_{t+1}^*}{\w_t^\top \widehat{\B} \w_t} \le  \min \left( {\sum_{i=2}^d \xi_{ti}^2 / \beta_i},\ {\xi_{t1}^2 / \beta_1} \right) \cdot  \frac{\left( \beta_1 - \beta_2 \right)^2}{32},
  \end{align}
  the algorithm converges linearly, and therefore we only need to solve $T = \calO \left( \log \frac{1}{\epsilon} \right)$ linear systems. But due to the zero initialization we use in the online setting, the ratio between initial error and final error for each $f_{t+1}$ is different from the offline setting.

  When $G(\rr_t) >1$, we are in the regime where ${\sum_{i=2}^d \xi_{ti}^2 / \beta_i} \ge {\xi_{t1}^2 / \beta_1}$, and we can ensure the sufficient accuracy in~\eqref{e:streaming-svrg-epsilon-t} by setting the ratio between the initial and the final error to be 
  \begin{align*}
    \eta_t = \frac{ (\beta_1 - \beta_2)^2 \left( \xi_{t1}^2 \right)}{ 16 \beta_1^2}
  \end{align*}
  in Lemma~\ref{lem:streaming-svrg-sample-complexity-for-f}. Since $\frac{\beta_1^2}{(\beta_1 - \beta_2)^2} \le 4$, this implies that
  \begin{align*}
    \frac{1}{\eta_t} \le \frac{64}{ \cos^2 \theta_t} = 64 ( 1 + \tan^2 \theta_t) 
    \le 64 \left( 1 + \frac{\beta_2}{\beta_1} G^2 (\rr_t) \right) \le 64 \left( 1 + G^2 (\rr_t) \right) 
\le  64 \left( 1 + G^2 (\rr_0) \right) .
  \end{align*}
Note that the sample complexity of this phase does not depend on the final accuracy in alignment.

  When $G(\rr_t) \le 1$, indicating that we are in the converging regime where ${\sum_{i=2}^d \xi_{ti}^2 / \beta_i} \le {\xi_{t1}^2 / \beta_1}$, we can ensure the sufficient accuracy in~\eqref{e:streaming-svrg-epsilon-t} by setting
  \begin{align*}
    \eta_t = \frac{ (\beta_1 - \beta_2)^2 \left( \sum_{i=2}^d \xi_{ti}^2 \right)}{ 16 \beta_1^2}
  \end{align*}
  in Lemma~\ref{lem:streaming-svrg-sample-complexity-for-f}. This implies that
  \begin{align*}
    \frac{1}{\eta_t} \le \frac{64}{ \sin^2 \theta_t}. 
  \end{align*}

  Our goal is to have $\sin^2 \theta_T \le \frac{\epsilon}{4}$, as this implies $\cos \theta_T = \sqrt{1 - \sin^2 \theta_T } \ge 1 - \sin^2 \theta_T \ge 1 - \frac{\epsilon}{4}$, and by Lemma~\ref{lem:online-transfer-align} this further implies $\text{align} \left( (\uu_T, \vv_T); (\uu^*, \vv^*) \right) \ge 1 - \epsilon$ as desired. 
  Since $\sin^2 \theta_t \le G^2 (\rr_t)$, and we have shown that $G^2 (\rr_t)$ decreases at a geometric rate, we can bound $\frac{1}{\sin^2 \theta_t}$ by a geometrically increasing series where the last term is $\frac{4}{\epsilon}$, and the sum of the truncated series up to time $T$ is of the same order of the last term, \ie, $\sum_{t=1}^T \frac{1}{\eta_t} = \calO \left( \frac{1}{\epsilon} \right)$. 

And the theorem follows from Lemma~\ref{lem:streaming-svrg-sample-complexity-for-f},  by summing the sample complexity of least squares problems over the outer shift-and-invert iterations. 

We remark that to achieve the result with probability $1 - \delta$, we require each least squares problem to be solved to the desired accuracy with failure probability $\delta/\log (1/\epsilon)$ (using the Markov inequality) and finally apply the union bound. This would only cause additional $\log (1/\epsilon)$ factors in the total sample complexity. 
\end{proof}

\section{SVRG with non-uniform sampling for finite-sum of nonconvex components}
\label{sec:nonuniform-svrg}

In this section, we show that for optimizing a convex objective that is
the finite-sum of nonconvex components, sampling each components with probability proportional to its
smoothness parameter, as shown in Algorithm~\ref{alg:nonuniform-svrg}, leads to improved convergence rate. In particular,
the final time complexity depends on average smoothness parameter rather than the
maximum smoothness.

\begin{algorithm}[t]
\caption{Non-uniform sampling SVRG for optimizing finite-sum of nonconvex
  components $F(\w)=\frac{1}{n} \sum_{i=1}^n f_i (\w)$.}
\label{alg:nonuniform-svrg}
  \renewcommand{\algorithmicrequire}{\textbf{Input:}}
  \renewcommand{\algorithmicensure}{\textbf{Output:}}
  \begin{algorithmic}
    \REQUIRE Stepsize $s$.
    \STATE Initialize $\w_0 \in \bbR^{d}$.
    \FOR{$j=1,2,\dots,M$}
    \STATE $\tu \leftarrow \w_{j-1}$
    \STATE Evaluate the batch gradient $\nabla F (\tu) = \frac{1}{n}
    \sum_{i=1}^n \nabla f_i (\tu)$
    \STATE $\uu_0 \leftarrow \w_{j-1}$
    \FOR{$t=1,2,\dots,m$}
    \STATE Randomly pick $i_t$ from $\{1,\dots,n\}$ with probability
    $\{p_i\}_{i=1}^n$. 
    \STATE $\uu_{t} \leftarrow \uu_{t-1} - s \left( \frac{\nabla f_{i_t}
      (\uu_{t-1}) -  \nabla f_{i_t} (\tu) }{p_{i_t} n} +  \nabla F (\tu) \right)$
    \ENDFOR
    \STATE $\w_{j} \leftarrow \frac{1}{n} \sum_{t=1}^m \w_{t}$
    \ENDFOR
    \ENSURE $\w_{M}$ is the approximate solution.
  \end{algorithmic}
\end{algorithm}

\begin{lemma}
Let $F(\w) = \frac{1}{n} \sum _{i=1}^n f_i (\w)$, where $F(\w)$ is
$\mu$-strongly convex, and  each component $f_i(\w)$ is $L_i$-smooth. Let
$\w^* = \argmin_{\w}\, F(\w)$.
In the inner loop of Algorithm~\ref{alg:nonuniform-svrg}, sample $i_t$ using weighted sampling probability
$\{p_i\}_{i=1}^n$ from $\{1,\dots,n\}$ where $p_i=\frac{L_i^2}{\sum_{j=1}^n
L_j^2}$, and set $s = \frac{2 \mu  n}{ 11 \sum_{i=1}^n L_i^2 }$, $m =
\frac{121 \sum_{i=1}^n L_i^2}{  8 n \mu^2 }$. 
Then the iteration complexity (number of vector operations) to reach
$\epsilon$-suboptimality is 
\begin{align*}
\calO \rbr{ \rbr{ n + \frac{\sum_{i=1}^n L_i^2}{n \mu^2} } \log
  \frac{(\frac{1}{n} \sum_{i=1}^n L_i) \cdot (F(\w_0) - F(\w^*))}{\mu \varepsilon} } .
\end{align*}
\end{lemma}
\begin{proof}
For the inner loop of Algorithm~\ref{alg:nonuniform-svrg}, we are performing updates of the following form:
\begin{align*}
\uu_{t} \leftarrow \uu_{t-1} - s \vv_t,
\end{align*} 
where
\begin{align*}
\vv_t = \frac{\nabla f_{i_t}(\uu_{t-1}) - \nabla f_{i_t}(\tu)}{p_{i_t} n} + \nabla F(\tu).
\end{align*}
Taking expectation over the random choice of component $i_t$, we have
\begin{align*}
\bbE_{t} [\vv_t] = \nabla F (\uu_{t-1})
\end{align*}
We now upper bound the variance of $\vv_t$:
\begin{align*}
\bbE_t \norm{ \vv_t - \nabla F(\uu_{t-1}) }^2 
= &  \bbE_t \sbr{ \frac{\nabla f_{i_t}(\uu_{t-1}) - \nabla f_{i_t}(\tu)}{p_{i_t} n} + \nabla F(\tu)  - \nabla F (\uu_{t-1})}^2 \\
= & \bbE_t \sbr{ \frac{1}{(p_{i_t} n)^2} \norm{ \nabla f_{i_t}(\uu_{t-1}) - \nabla f_{i_t}(\tu) }^2  }
 - \norm{ \nabla F(\uu_{t-1}) - \nabla F(\tu) }^2 \\
\le & \frac{1}{n} \sum_{i=1}^n \frac{1}{p_{i} n}  \norm{  \nabla  f_{i}(\uu_{t-1}) - \nabla f_{i}(\tu) }^2 \\
\le & \frac{2}{n} \sum_{i=1}^n \frac{1}{p_{i} n} \rbr{ \norm{ \nabla
   f_{i}(\uu_{t-1}) - \nabla f_{i}(\w^*) }^2  + \norm{ \nabla f_{i}(\tu) - \nabla f_{i}(\w^*) }^2 }\\
\leq& \frac{2}{n} \sum_{i=1}^n \frac{L_i^2}{p_i n} \rbr{ \norm{\uu_{t-1} - \w^*}^2 + \norm{\tu - \w^*}^2 }
\end{align*}
where we have used the fact that $\bbE \norm{\x - \bbE[\x]}^2 = \bbE
\norm{\x}^2 - (\bbE[\x])^2$ for a random vector $\x$ in the second
equality, and that $\norm{\x+\y}^2 \le 2 \norm{\x}^2 + 2 \norm{\y}^2$ in
the second inequality. 

By choosing
\begin{align*}
p_i = \frac{L_i^2}{\sum_{i=1}^n L_i^2},
\end{align*}
the above inequality becomes
\begin{align*}
\bbE_t \norm{ \vv_t - \nabla F(\uu_{t-1}) }^2  \leq \frac{2 \sum_{i=1}^n L_i^2 }{n} \rbr{ \norm{\uu_{t-1} - \w^*}^2 + \norm{\tu - \w^*}^2 }.
\end{align*}

Define $\bar{L}=\frac{1}{n} \sum_{i=1}^n L_i$ which is an upper bound of the smoothness
parameter of the average function $F(\w)$ as
\begin{align*}
F(\aa) - F(\b) & = \frac{1}{n} \sum_{i=1}^n {f_i (\aa) - f_i (\b)} \le
\frac{1}{n} \sum_{i=1}^n \inner{\nabla f_i (\b)}{\aa-\b} +\frac{L_i}{2}
\norm{\aa-\b}^2 \\
& = \inner{\nabla F(\b)}{\aa-\b} + \frac{\frac{1}{n} \sum_{i=1}^n L_i}{2} \norm{\aa-\b}^2,
\end{align*}
and define $\hat{L}=\sqrt{\frac{1}{n} \sum_{i=1}^n L_i^2}$. 
We then bound the distance from each iterate to the optimum:
\begin{align*}
& \bbE_t \norm{\uu_t - \w^*}^2 = \norm{\uu_{t-1} - \w^*}^2 - 2 s \inner{\uu_{t-1} - \w^*}{\bbE_t \sbr{ \vv_t } }  + s^2 \bbE_t \norm{\vv_t}^2 \\
\leq& \norm{\uu_{t-1} - \w^*}^2 - 2 s \inner{\uu_{t-1} - \w^*}{\nabla  F(\uu_{t-1})} + s^2 \norm{\nabla F(\uu_{t-1})}^2
+ 2 s^2 \hat{L}^2 \rbr{ \norm{\uu_{t-1} - \w^*}^2 + \norm{\tu - \w^*}^2 } \\
\leq& \norm{\uu_{t-1} - \w^*}^2 - 2 s \mu \norm{\uu_{t-1} - \w^* }^2 + s^2 \bar{L}^2 \norm{\uu_{t-1} - \w^*}^2
+ 2 s^2 \hat{L}^2 \rbr{ \norm{\uu_{t-1} - \w^*}^2 + \norm{\tu - \w^*}^2 }
\end{align*}
where we have used the fact that $\bbE \norm{\x} = (\bbE[\x])^2 + \bbE
\norm{\x - \bbE[\x]}^2$ in the first inequality, and the smoothness and
strong convexity of $F(\w)$ in the second inequality.

By the Jensen's inequality, we have $\bar{L} \le \hat{L}$. 
Therefore, we continue from above and obtain
\begin{align*}
\bbE_t \sbr{ \norm{\uu_t - \w^*}^2 } - \bbE \sbr{ \norm{\uu_{t-1} - \w^*}^2 }  \leq \rbr{ 
3 s^2 \hat{L}^2 - 2 s \mu} \norm{ \uu_{t-1} - \w^* }^2
  + 2 s^2 \hat{L}^2 \norm{\tu - \w^*}^2 .
\end{align*}
Summing the above inequality over the inner loop yields
\begin{align*}
\bbE \norm{\uu_m - \w^*}^2 - \bbE \norm{\uu_0 - \w^*}^2 \leq \rbr{ 
3 s^2 \hat{L}^2 - 2 s \mu} \sum_{t=1}^m  \norm{ \uu_{t-1} - \w^* }^2  + 2
  m  s^2 \hat{L}^2 \norm{\tu - \w^*}^2
\end{align*}
Using $\uu_0 = \tu$ and rearranging terms, we have
\begin{align*}
\rbr{ 2 s \mu - 3 s^2 \hat{L}^2 }  \sum_{t=1}^m  \norm{ \uu_{t-1} - \w^* }^2 \leq \rbr{1 + 2 m s^2 \hat{L}^2} \norm{\tu - \w^*}^2
\end{align*}
Using $\tu = \w_{j-1}$ and $\w_{j} = \frac{1}{m} \sum_{t=0}^{m-1} \uu_{t}$, we obtain
\begin{align*}
\bbE \norm{ \w_{j} - \w^*}^2 \leq \frac{1 + 2 m s^2 \hat{L}^2}{  2 m s \mu  - 3 m  s^2 \hat{L}^2 }  \bbE \norm{ \w_{j-1} - \w^*}^2.
\end{align*}
Setting
\begin{align*}
s = \frac{2 \mu }{ 11 \hat{L}^2}, \qquad m = \frac{1}{ 2 s^2 \hat{L}^2} =
  \frac{121 \hat{L}^2}{8\mu^2},
\end{align*}
we obtain
\begin{align*}
\bbE \norm{ \w_{j} - \w^*}^2 \leq \frac{1}{2}\bbE \norm{ \w_{j-1} - \w^*}^2.
\end{align*}
Therefore the squared distance to minimum decreases geometrically for
each outer loop. After $M$ iterations, we have
\begin{align*}
F(\w_M) - F(\w^*) \le \frac{\bar{L}}{2} \norm{\w_M - \w^*}^2 \le
  \frac{\bar{L}}{2}  \left(\frac{1}{2}\right)^M \norm{\w_0 - \w^*}^2 \le
  \frac{\bar{L}}{\mu} \left(\frac{1}{2}\right)^M  (F(\w_0) - F(\w^*)).
\end{align*}
Setting the right hand side to $\varepsilon$ gives the number of outer
iterations $M = \calO \rbr{ \log \frac{(\bar{L}/\mu) \cdot (F(\w_0) - F(\w^*))}{\varepsilon}}$. 
Finally, the total iteration complexity to reach $\varepsilon$-suboptimality is
\begin{align*}
\calO \rbr{ (n + m) M} = \calO \rbr{ \rbr{ n + \frac{\hat{L}^2}{\mu^2} }
  \log \frac{\bar{L} (F(\w_0) - F(\w^*))}{\mu \varepsilon}  }.
\end{align*}
\end{proof}
\end{document}